\documentclass[letterpaper, 10pt, conference]{ieeeconf}
\IEEEoverridecommandlockouts
\usepackage{epsfig}
\usepackage{amssymb}
\usepackage{cite}
\usepackage{booktabs}
\usepackage{multirow}
\usepackage{subcaption}
\usepackage{mathtools}
\usepackage{siunitx}
\usepackage{comment}
\usepackage{threeparttable}
\usepackage{bm}
\usepackage{color}

\newcommand{\figlab}[1]{\label{fig:#1}}
\newcommand{\figref}[1]{Fig.~\ref{fig:#1}} % Figure
\newcommand{\tablab}[1]{\label{tab:#1}}
\newcommand{\tabref}[1]{Table~\ref{tab:#1}} % Table
\newcommand{\forlab}[1]{\label{for:#1}}
\newcommand{\forref}[1]{Equation~(\ref{for:#1})} % Equation
\newcommand{\algolab}[1]{\label{algorithm:#1}}
\newcommand{\algoref}[1]{Algorithm~\ref{algorithm:#1}} % Equation

\newcommand{\etal}{\textit{et~al.}}
\newcommand{\ie}{\textit{i.e.,}}
\newcommand{\eg}{\textit{e.g.,}}
\usepackage{amssymb}% http://ctan.org/pkg/amssymb
\usepackage{pifont}% http://ctan.org/pkg/pifont
\newcommand{\cmark}{\ding{51}}%
\newcommand{\xmark}{\ding{55}}%

\usepackage{amsmath}
\usepackage{algorithmicx}
\usepackage[ruled]{algorithm}
\usepackage[noend]{algpseudocode}

\usepackage{array}
\title{\LARGE \textbf{Soft-Jig-Driven Assembly Operations}}
\author{Takuya Kiyokawa, Tatsuya Sakuma, Jun Takamatsu, and Tsukasa Ogasawara%
\thanks{All authors are with the Division of Information Science, Robotics Laboratory, Nara Institute of Science and Technology (NAIST), Japan {\tt\small \{kiyokawa.takuya.kj5, sakuma.tatsuya.sn1, j-taka, ogasawar\}}@is.naist.jp}}

\begin{document}

\maketitle
\thispagestyle{empty}
\pagestyle{empty}

\begin{abstract}
To design a general-purpose assembly robot system that can handle objects of various shapes, we propose a soft jig that fits to the shapes of assembly parts. The functionality of the soft jig is based on a jamming gripper developed in the field of soft robotics.
The soft jig has a bag covered with a malleable silicone membrane, which has high friction, elongation, and contraction rates for keeping parts fixed. The bag is filled with glass beads to achieve a jamming transition.
We propose a method to configure parts-fixing on the soft jig based on contact relations, reachable directions, and the center of gravity of the parts that are fixed on the jig. 
The usability of the soft jig was evaluated in terms of the fixing performance and versatility for various shapes and postures of parts. 
\end{abstract}

\section{Introduction} \label{intro}
There is an ever-increasing demand for an agile manufacturing system~\cite{Gunasekaran1999,Costa2017} that can flexibly respond to the variety of products required by markets worldwide.
To achieve high-mix low-volume production in the manufacturing industry, an assembly robot with high versatility, as proposed in~\cite{Maeda2003,Kim2019,Gorjup2020}, is required to operate various mechanical parts.

Generally, jigs are used to efficiently assemble different types of products~\cite{Rajan1999,Zhang2019} for mass production.
However, in high-mix low-volume production, it is impractical to develop custom-made jigs every time a product is replaced.

In this study, we develop a deformable fixing device named \textit{soft jig}, as shown in~(a) and (b) of \figref{device}. 
A soft jig is highly versatile as a fixture for different parts with various shapes as the jig surface deforms according to the shapes.
The fixing ability of the soft jig based on a jamming transition can be used to hold assembly parts by creating datum planes (\figref{device}(c)) on the jig surface (\figref{device}(d)). 
%%%%%%%%%%%%%%%%%%%%%%%%%%%%%%%%%%
\begin{figure}[tb]
  \centering
  {\begin{minipage}[tb]{0.40\linewidth}
    \includegraphics[width=0.85\linewidth]{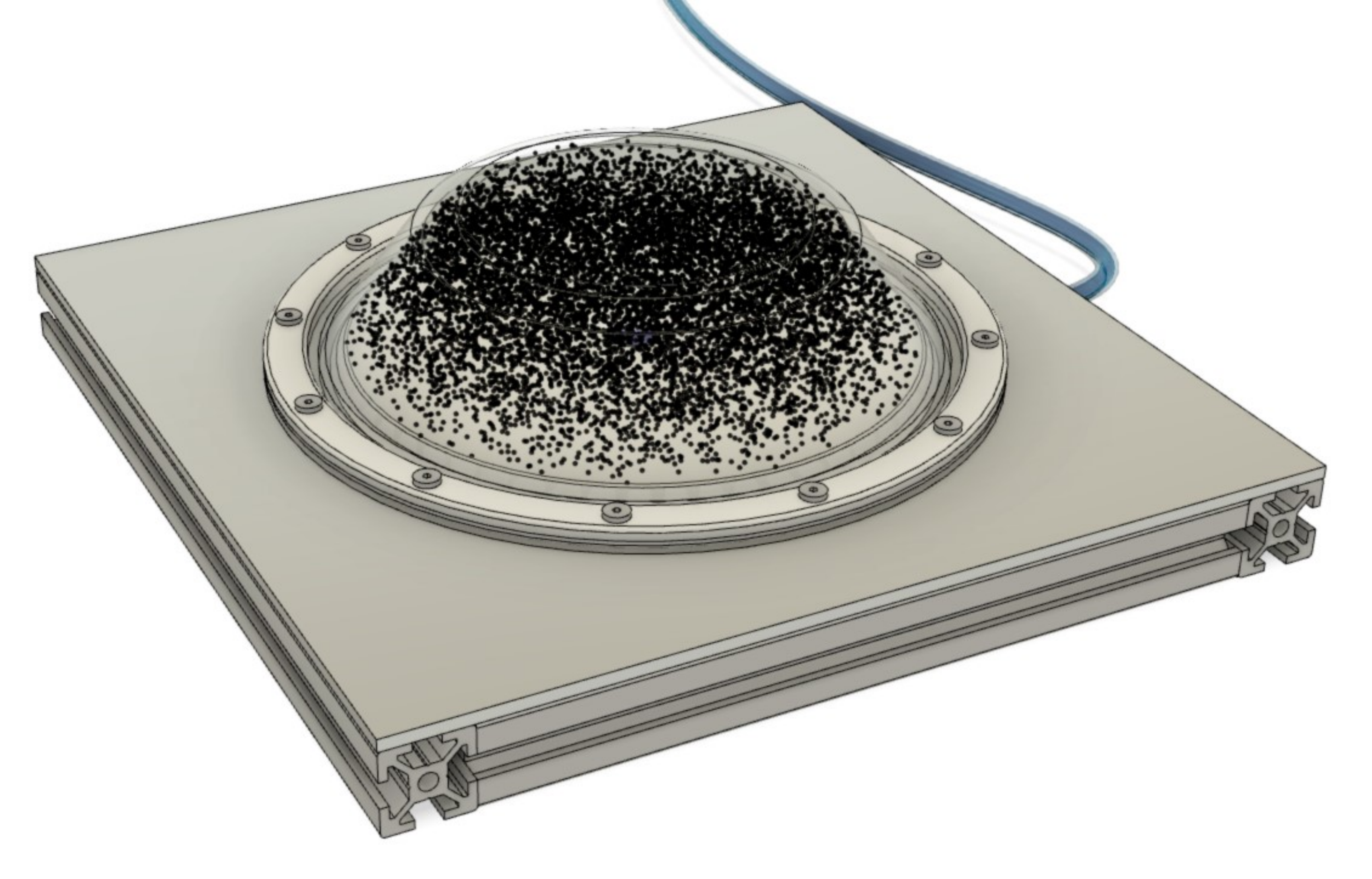}
    \subcaption{\small{Appearance}}\figlab{appearance}
  \end{minipage}
  \begin{minipage}[tb]{0.58\linewidth}
    \includegraphics[width=0.9\linewidth]{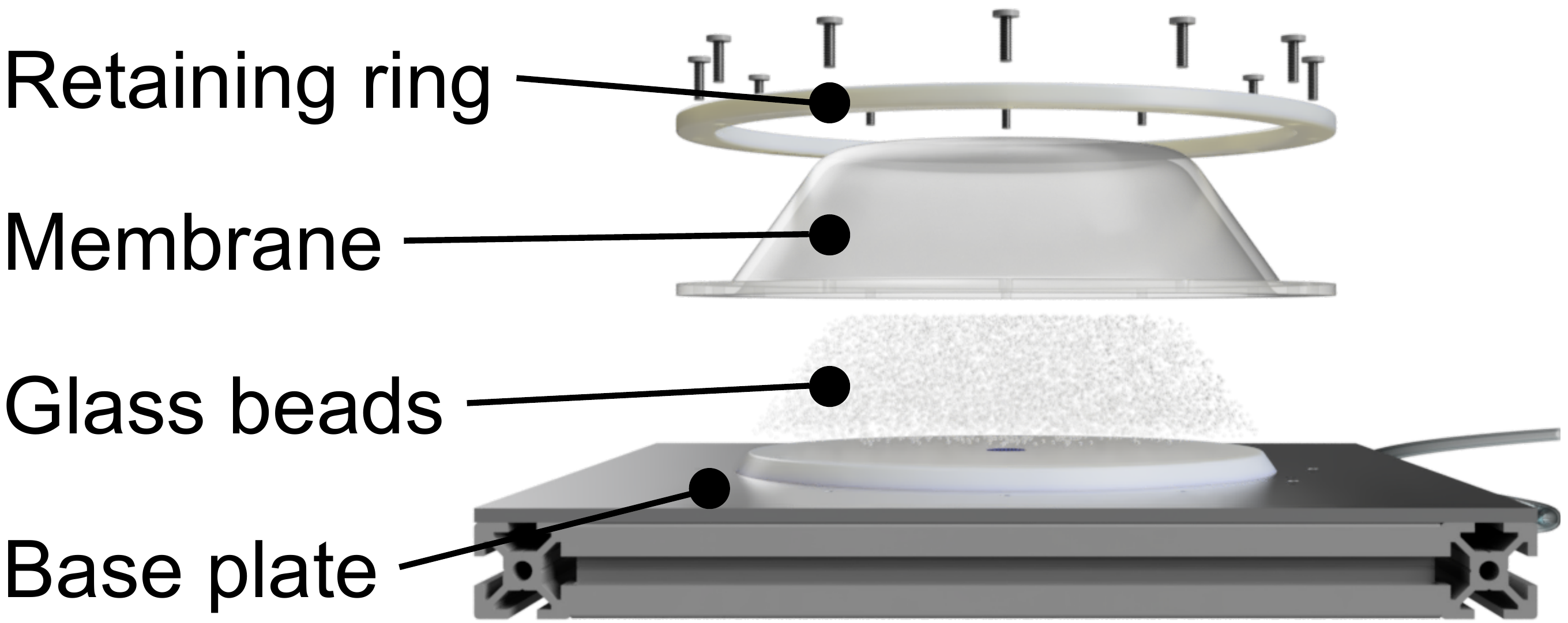}
    \subcaption{\small{Exploded view}}\figlab{exploded}
  \end{minipage}
  \begin{minipage}[tb]{0.49\linewidth}
    \includegraphics[width=0.82\linewidth]{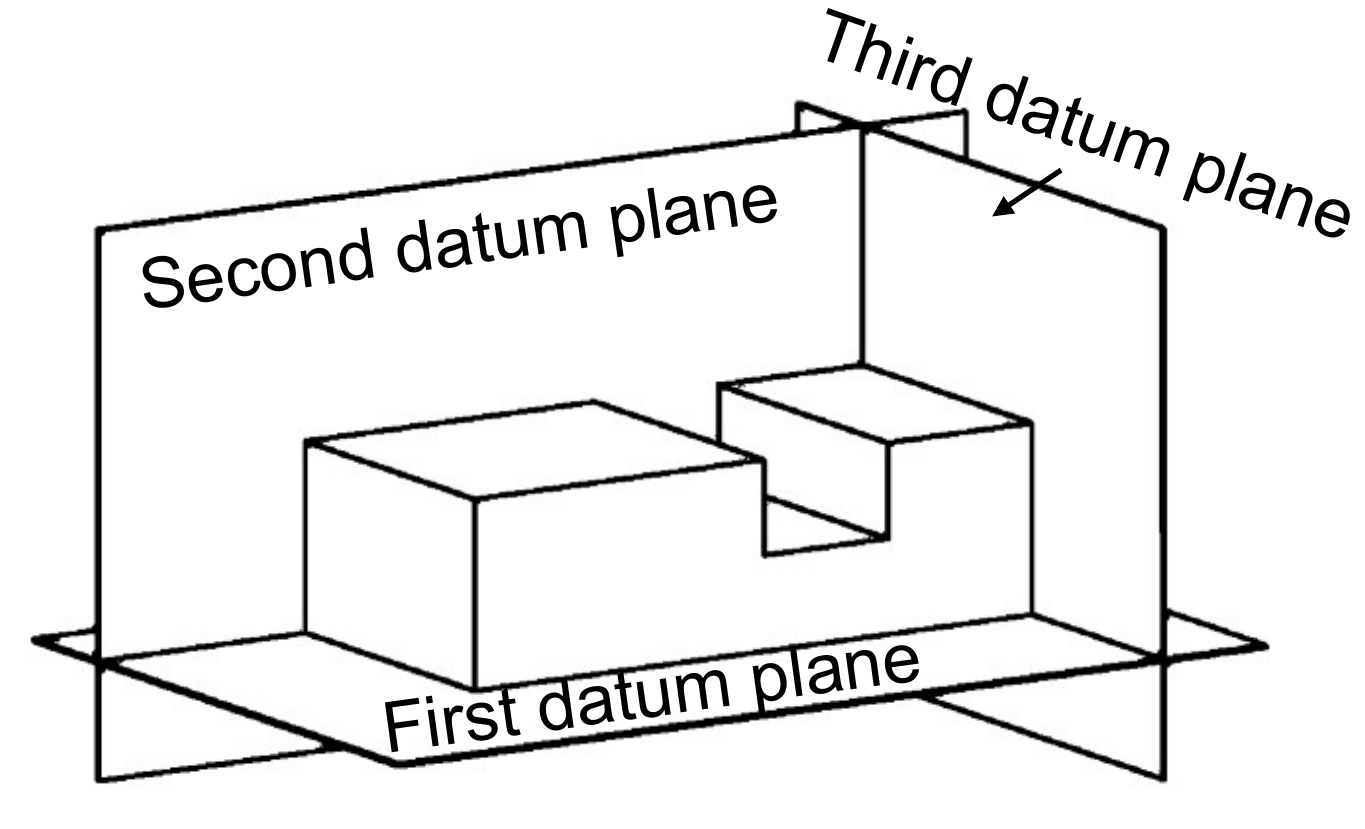}
    \subcaption{\small{Datum planes~\cite{Trappey2005}}}\figlab{datum-planes}
  \end{minipage}
  \begin{minipage}[tb]{0.49\linewidth}
    \includegraphics[width=0.8\linewidth]{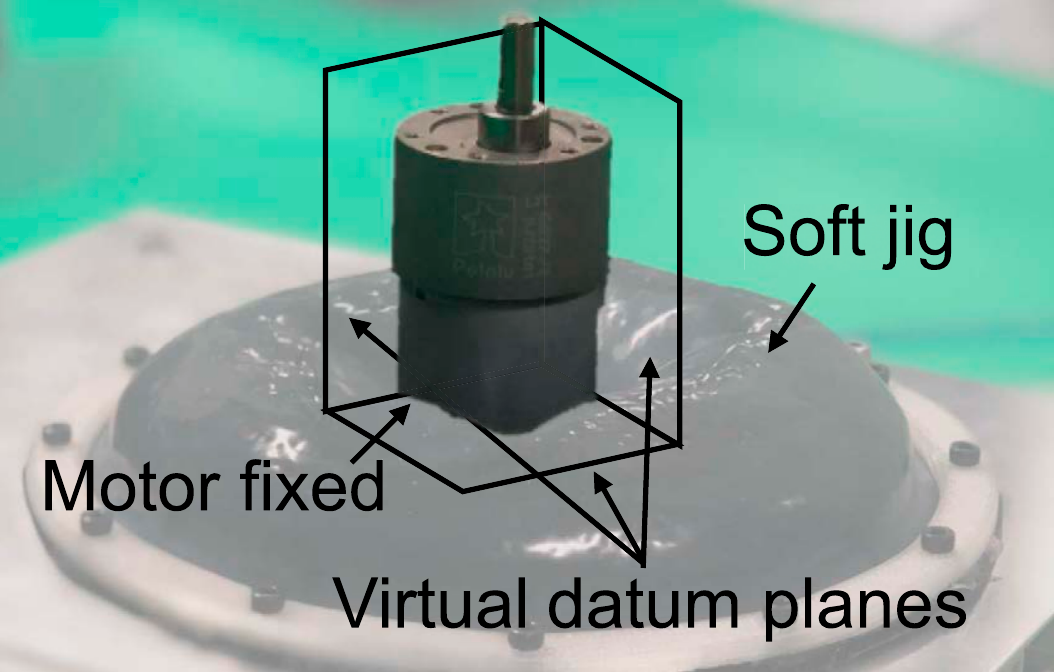}
    \subcaption{\small{Creation of datum planes}}\figlab{datum-planes-created}
  \end{minipage}}
  \caption{\small{Design of soft jig. The appearance (a) and exploded view (b) of the soft jig are shown. Datum planes (c) (this figure is created with reference to Fig.~2 in~\cite{Trappey2005}) are needed to fix objects in a certain pose, and they are created on the malleable membrane (d).}}\figlab{device}
\end{figure}
%%%%%%%%%%%%%%%%%%%%%%%%%%%%%%%%%%
\begin{figure*}[t]
    \centering
    \includegraphics[width=0.75\linewidth]{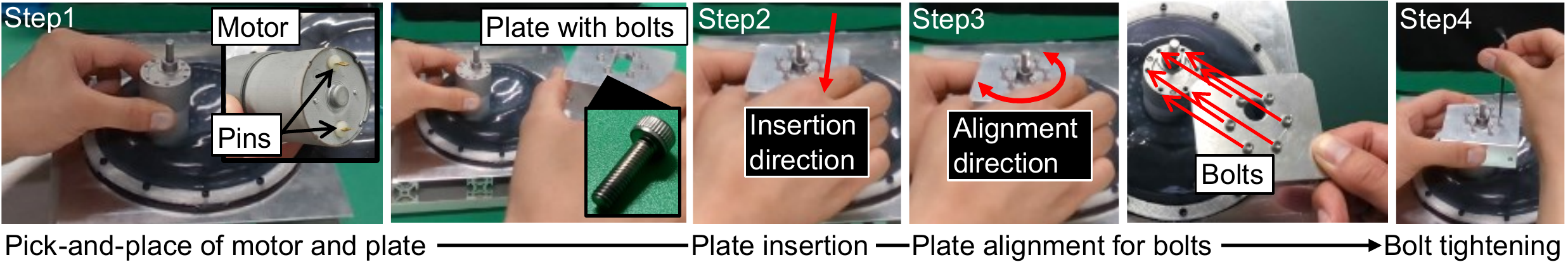}
    \caption{\small{Manual assembly sequence with the soft jig. In the pick-and-place of the motor, we consider if the plate can be placed onto the soft jig. Consequently, we can fix a motor even if the bottom surface of the motor has pins because of the malleable membrane.}}
    \figlab{human-insertion}
\end{figure*}
%%%%%%%%%%%%%%%%%%%%%%%%%%%%%%%%%%

The primary objective of this study is to provide a new parts-fixing device for robotic assembly. 
Specifically, the proposed soft jig provides a new concept for general-purpose assembly jigs that can be used for a flexible assembly robot system. 
We propose a design of the soft jig capable of utilizing the jamming transition as the fixing mechanism.

In parts-assembly with the soft jig, we determine fixed parts and their postures based on the three requirements: (1) to contact between parts in one assembly operation (\eg~placement and insertion), ignoring the parts without any contacts (2) to ensure a fixed part has reachable directions to the parts assembled on it, selecting the fixed part and posture by extracting interference-free parts-displacement directions using CAD, and (3) to develop a part posture with a low center of gravity (CoG) to achieve mechanical stability, selecting the fixed posture based on CAD-based calculations of CoGs for all candidates of the fixed parts.

Our experiments demonstrate the fixing performance and versatility of the soft jig for different fixing configurations. 
Further,  we examine the feasibility of assembly operations using an actual robot and discuss parts-pose estimation.

\section{Related Work}
Several jig-less operation methods~\cite{Naing2000,Kim2017} and general-purpose jig-designing methods~\cite{Bi2001} have been proposed to reduce the human effort for designing custom-made jigs.
Several researchers~\cite{Grippo1987,Whybrew1992,Fathianathan2007} developed systems to automatically design rigid modular fixtures by combining elements such as locators, blocks, and clamps.

Although high-precision positioning is possible with rigid jigs, they must be replaced to better correspond to the fixed parts with different shapes.
Furthermore, previous approaches to substitute the rigid jigs needed control of the actuators~\cite{Naing2000,Kim2017,Shi2020} or calculations for the shape optimization~\cite{Grippo1987,Whybrew1992,Bi2001}.
Hence, their versatility and ease of fixing are relatively low.
\tabref{pros-cons} presents a comparison between the previous and proposed general-purpose assembly jigs. 
The proposed fixture requires a pose estimation whereas increasing the easiness of fixing and the versatility.
%%%%%%%%%%%%%%%%%%%%%%%%%%%%%%%%%%
\begin{table}[t]
    \centering
        \caption{\small{Comparison between previous and proposed general-purpose assembly jigs.}}
        \tablab{pros-cons}
        \begin{tabular}{p{7mm}p{9mm}p{9mm}p{9mm}} \toprule
            \multicolumn{1}{c}{Method} & \multicolumn{1}{c}{Easiness of fixing} & \multicolumn{1}{c}{Versatility} & \multicolumn{1}{c}{Parts-positioning} \\ \midrule
            \multicolumn{1}{l}{Jig with supports} & \multicolumn{1}{c}{\begin{tabular}{c}\xmark$^{\rm *a}$\end{tabular}} & \multicolumn{1}{c}{\begin{tabular}{c}\cmark$^{\rm *b}$\end{tabular}} &
            \multicolumn{1}{c}{\begin{tabular}{c}\cmark$^{\rm *c}$\end{tabular}} \rule[-1mm]{0mm}{4mm} \\
            \multicolumn{1}{l}{Soft jig (proposed)} & \multicolumn{1}{c}{\begin{tabular}{c}\cmark$^{\rm *d}$\end{tabular}} & \multicolumn{1}{c}{\begin{tabular}{c}\cmark$^{\rm *e}$\end{tabular}} & 
            \multicolumn{1}{c}{\begin{tabular}{c}\xmark$^{\rm *f}$\end{tabular}} \rule[-1mm]{0mm}{4mm} \\ 
            \bottomrule
        \end{tabular}
        \begin{tablenotes}
          \item[a]\footnotesize{$^{\rm *a}$ Supports need to be placed in postures}
          \item[b]\footnotesize{$^{\rm *b}$ Surrounding supports fit the object shape}
          \item[c]\footnotesize{$^{\rm *c}$ Rigid supports fix the object in a certain pose}
          \item[d]\footnotesize{$^{\rm *d}$ On-off control of air pressure}
          \item[e]\footnotesize{$^{\rm *e}$ Deformable body fits the object shape}
          \item[f]\footnotesize{$^{\rm *f}$ Stiffness of the malleable membrane is lower than that of the metal jig surface}
        \end{tablenotes}
\end{table}
%%%%%%%%%%%%%%%%%%%%%%%%%%%%%%%%%%

Several studies have been conducted using flexible robotic end-effectors that fit the object shapes that need to be manipulated~\cite{Lee2005,Watanabe2017}.
Brown~\etal~\cite{Brown2010} proposed a jamming gripper that can grasp rigid objects of various shapes.
The gripper surface is covered with a silicon membrane filled with powder particles. 
The extensibility of the jamming gripper has been discussed in terms of the parts-recognition~\cite{Sakuma2018,Alspach2019} and sensing for robot manipulation~\cite{Sakuma2019,Lu2020}. 

The applicability of the jamming gripper has been researched for different purposes, such as in the feet of robots for walking on natural terrain~\cite{Lathrop2020,Chopra2020} and climbing walls~\cite{Fujita2018}. 
Such soft robotics technologies are expected to be applied to the field of robotic assembly~\cite{HamayaIROS,HamayaICRA} for high-mix low-volume production.

\section{Assumptions and Problem Setting} \label{assum}
The fixing planning and shape of custom-made jigs for mass production are designed according to an assembly sequence. 
In contrast, the fixing planning of the soft jig for high-mix low-volume production needs to be considered independently from the assembly sequence of the short life-cycle products. 
Thus, a certain assembly sequence is given.

As an example, let us consider the assembly task shown in~\figref{human-insertion}; three types of parts shown in~\figref{parts}(a) are handled. 
The assembly task includes fundamental operations frequently conducted by humans: picking, placing, inserting, and screwing~\cite{Yamazaki2018,Fukuda2019}.
First, we grasp the motor and insert the motor shaft into the plate's hole. Subsequently, we align the bolts with the motor's holes and tighten the bolts.

One assembly step consists of assembling two (partially assembled) parts; thus, the parts that should be in contact with other parts are target parts for the assembly with the soft jig. 
One way to achieve this is to fix one part and manipulate the other part. 
As the manipulated part needs to be reachable from the outside to connect to the fixed part, the fixed pose must be planned as interference-free.
At least one manipulator should manipulate a part on the part stably fixed. 
Thus, we make the CoG of the fixed part low.
%%%%%%%%%%%%%%%%%%%%%%%%%%%%%%%%%%
\begin{figure}[tb]
  \centering
  {\begin{minipage}[tb]{0.49\linewidth}
    \centering
    \includegraphics[width=0.74\linewidth]{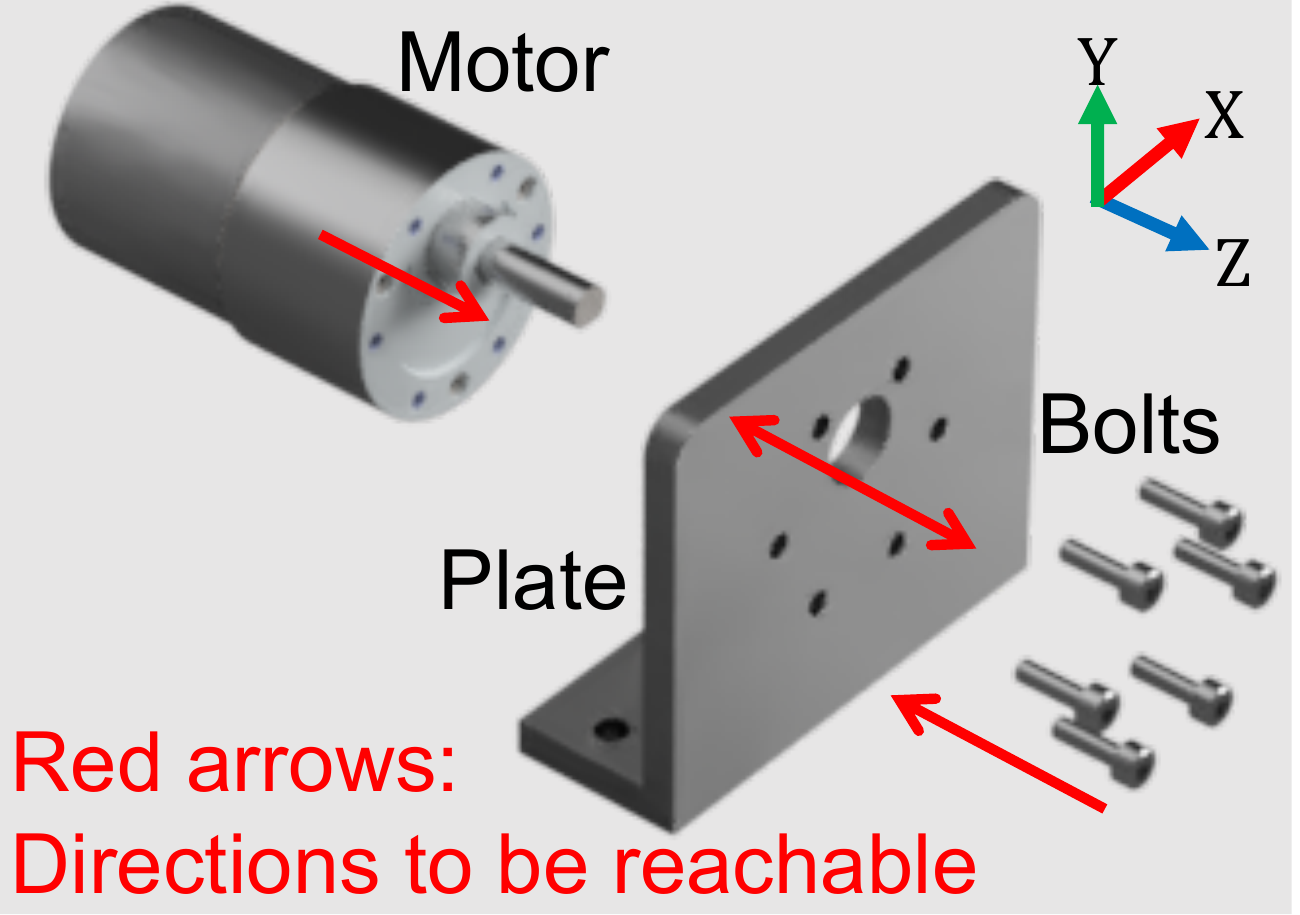}
    \vspace{0.5mm}
    \subcaption{\small{Structure of model}}\figlab{disassembled-parts}
  \end{minipage}
  \begin{minipage}[tb]{0.49\linewidth}
    \centering
    \includegraphics[width=0.76\linewidth]{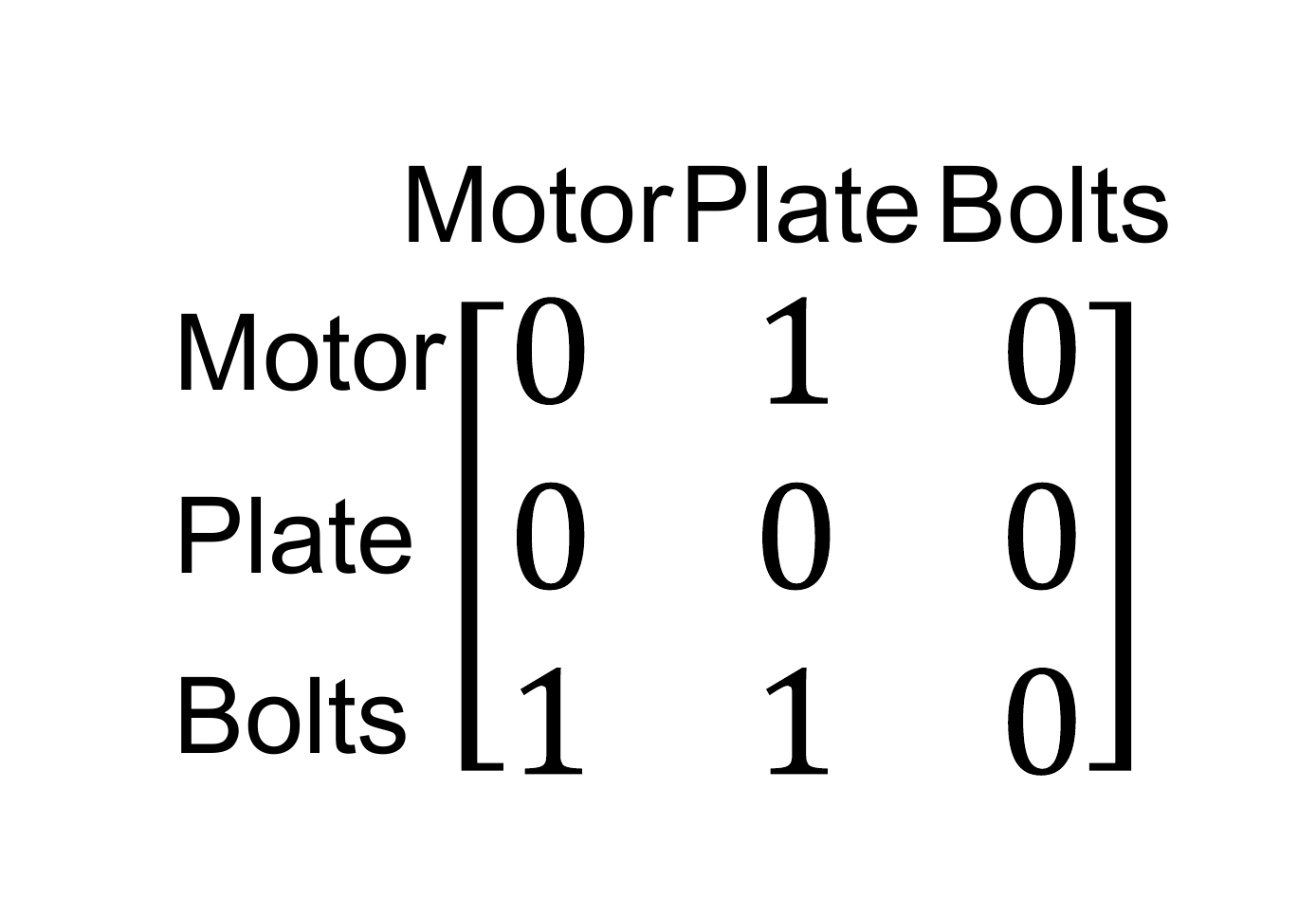}
    \subcaption{\small{Contact matrix}}\figlab{insertion-matrix}
  \end{minipage}
  \begin{minipage}[tb]{\linewidth}
    \centering
    \includegraphics[width=0.76\linewidth]{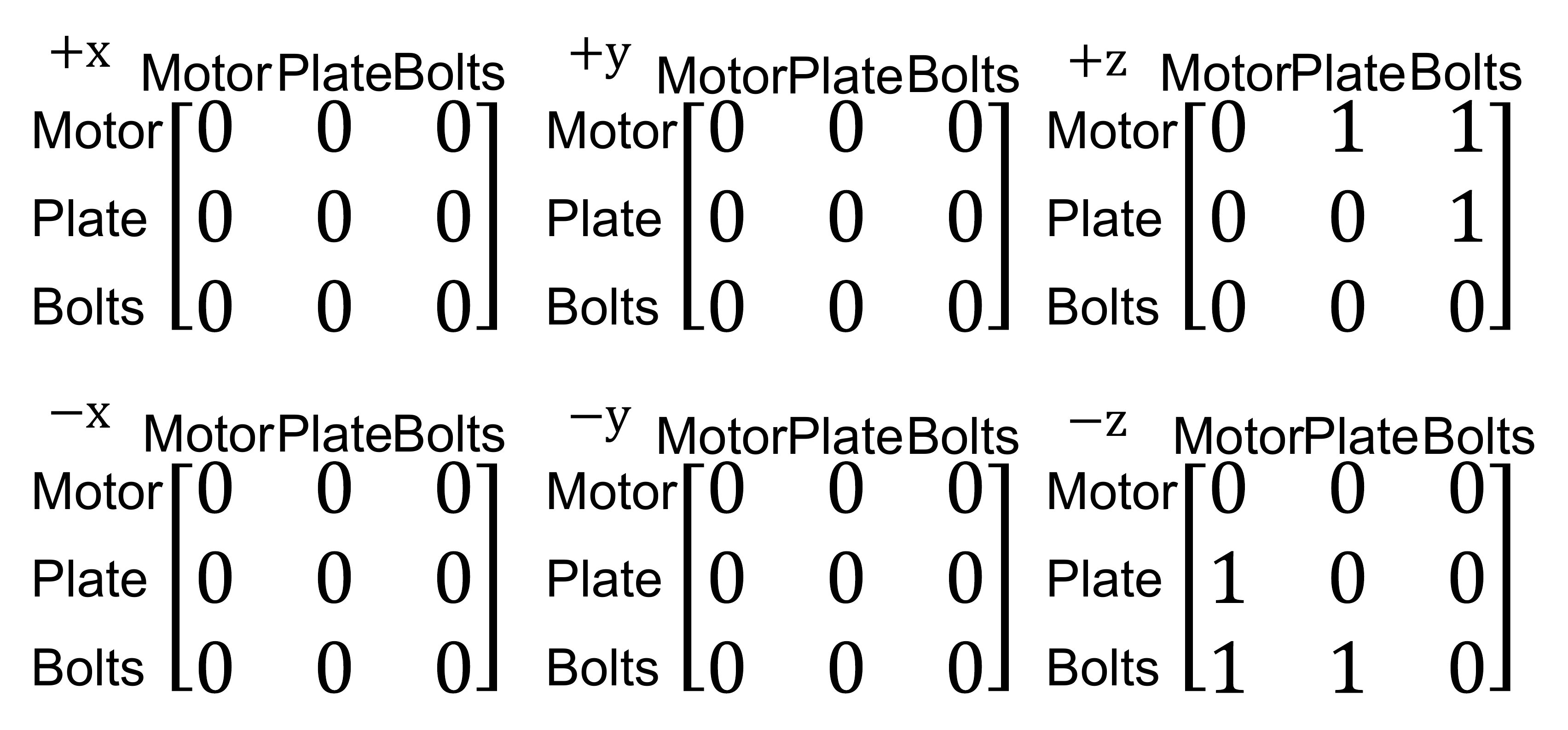}
    \subcaption{\small{Interference-free matrix}}\figlab{interference-matrix}
  \end{minipage}}
  \caption{\small{Assembly parts, contact matrix, and interference-free matrix used in our experiments. The two matrices are calculated based on an assembled CAD model and are used to configure the parts-fixing.}}\figlab{parts}
\end{figure}
%%%%%%%%%%%%%%%%%%%%%%%%%%%%%%%%%%

\section{Design of Soft Jig} \label{jig}
%%%%%%%%%%%%%%%%%%%%%%%%%%%%%%%%%
\begin{figure}[tb]
  \centering
  \includegraphics[width=0.95\linewidth]{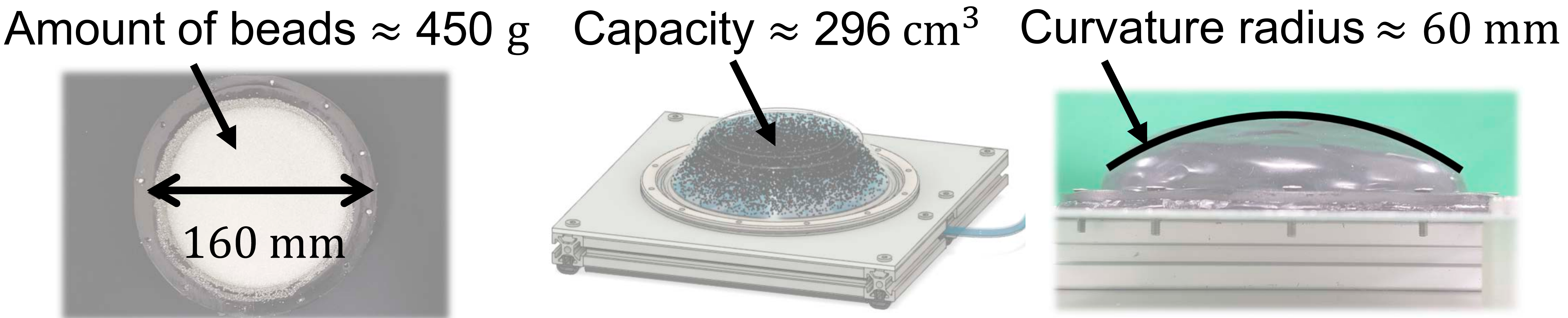}
  \caption{\small{Specifications of soft jig.}}\figlab{detail-geo}
\end{figure}
%%%%%%%%%%%%%%%%%%%%%%%%%%%%%%%%%%
The designed structure of the soft jig is depicted in (a) and (b) of \figref{device}. \figref{detail-geo} shows the specifications of the proposed soft jig.
Silicone rubber with a Shore A hardness of 2 (Smooth-On, Dragon Skin FX-Pro) was used for the elastomer membrane (\SI{1}{mm} in thickness and \SI{160}{mm} in diameter) to form the malleable surface of a bag. 

The bag with \SI{296}{cm^3} capacity was filled with glass beads of \SI{450}{g} with a diameter of approximately \SI{1}{mm} (Fuji Manufacturing Co., Ltd., Fuji Glass Beads FGB-20). 
We selected the glass beads because they do not corrode. The curvature radius of the bag was approximately~\SI{60}{mm}. 

The rigidity of the soft jig can be altered by vacuuming out the air inside the jig through an air port under the jig base, and a target part can be fixed.
We use an off-board vacuum pump, and the confining pressure inside the bag was approximately \SI{90}{kPa}.
The high friction, elongation, and contraction ratio of the elastomer material are beneficial to the parts-fixing performance.

The parts-fixing process is as follows: before placing the target part, the jig surface is initialized by pumping air into the jig because the fixing performance depends on the initial shape of the surface~\cite{Amend2012}.
Subsequently, the part is grasped, transported, and placed on the jig.
When the part is pushed onto the jig, it becomes fixed by taking advantage of the jamming transition by evacuating air from inside the jig.

\section{Configuring Parts-Fixing} \label{config}
The proposed parts-fixing algorithm (\algoref{fixing}) is based on the aformentioned three requirements (Section \ref{intro}) and assumptions (Section \ref{assum}). 
Given an assembly sequence, we decide which assembly part to place in which pose.
Specifically, the proposed algorithm selects a fixed part and a fixing posture that allows other objects to reach the fixed part. 
In addition, the CoG position of the fixed part is low. 
%%%%%%%%%%%%%%%%%%%%%%%%%%%%%%%%%%%%%%%%%%%%%%%%%%%%%%%%%%%%
\alglanguage{pseudocode}
\begin{algorithm}[t]
\caption{Parts-Fixing Configuration Algorithm} \algolab{fixing}
\begin{algorithmic}[1]
\renewcommand{\algorithmicrequire}{\textbf{Input:}}
\renewcommand{\algorithmicensure}{\textbf{Output:}}
\Require Assembly order $\{P_1,P_2,..,P_{\eta}\}$
\Ensure Parts $\hat{\bm{P}}^*$ to be fixed in certain postures $\hat{\bm{A}}^*$
\Procedure{CONFIGURE-FIXING-PARTS}{}
\State $P_t \gets P_1$
\State Set $\bm{A}_{det}$, $\hat{\bm{P}}^*$, and $\hat{\bm{A}}^*$ to empty lists
\For {$i=2,\ldots,\eta$}
	\State Calculate $\bm{A}(P_t,P_i)$ using \forref{place-determine}
    \If {Elements of $\bm{A}(P_t,P_i)$ are all 0}
        \State \textbf{break}
    \EndIf
    \State Generate model $P_c$ by combining $P_t$ and $P_i$
    \State $P_t \gets$ combined part $P_c$
    \State $\bm{A}_{det} \gets$ determined postures $\bm{A}(P_t,P_i)$
    \If{$\bm{A}_{det}$ includes two or more 1}
        \State Calculate CoG of model $P_{t}$ using CAD
        \State Determine posture $\hat{A}$ based on the CoG
    \Else
        \State $\hat{A} \gets$ the first element of $\bm{A}_{det}$
    \EndIf
    \State Set $\hat{P}$ to the bottom part in posture $\hat{A}$
    \State Add $\hat{P}$ and $\hat{A}$ to $\hat{\bm{P}}^*$ and $\hat{\bm{A}}^*$, respectively
\EndFor
\EndProcedure
\end{algorithmic}
\end{algorithm}
%%%%%%%%%%%%%%%%%%%%%%%%%%%%%%%%%%%%%%%%%%%%%%%%%%%%%%%%%%%%

Given an assembly order \{$P_1,P_2,..,P_{\eta}$\}, we can obtain a list, $\hat{\bm{P}}^*$, of the fixed parts and a list, $\hat{\bm{A}}^*$, of the fixed postures.
The process starts by initializing a target part, $P_t$, as $P_1$.
The determined posture list $\bm{A}_{det}$, fixed parts list $\hat{\bm{P}}^*$, and corresponding fixed postures list $\hat{\bm{A}}^*$ are initialized as empty lists.

In the main routine, we first calculate a reachable direction list, $\bm{A}(P_i,P_k)$.
To perform interference-free operations, we calculate the reachable direction matrix $\bm{W}_j~(j \in \{+x, -x, +y, -y, +z, -z\})$, with contact matrix $\bm{C}$~\cite{Bedeoui2019} and interference-free matrix $\bm{M}_j$~\cite{Tariki2020,TarikiAR} shown in~\figref{parts}(b) and (c).
The reachable direction matrix, in which the element reachable from the $j$ direction is 1, is written as
\begin{equation}\forlab{reachable}
\bm{W}_j = (\bm{C} + \bm{C}^{\mathrm{T}}) \odot \bm{M}_j,
\end{equation}
where $\odot$ is the Hadamard product of matrices.

Here, the parts are expressed as $P_1,P_2,..,P_\eta$ (where $\eta$ is the number of parts). 
The reachable direction list, $\bm{A}(P_i,P_k)$, with regard to the translational displacement is calculated as
\begin{equation}\forlab{place-determine}
 (W_{+x}(P_i,P_k),
  W_{-x}(P_i,P_k),
  \hdots,
  W_{-z}(P_i,P_k)).
\end{equation}

If $\bm{A}(P_i,P_k)$ is a list filled with 0, the process is ended.
Otherwise, our method generates model $P_c$ by combining $P_t$ and $P_i$. 
The next process consists of updating $P_t$ by $P_c$ and substituting the determined postures, $\bm{A}(P_i,P_k)$, for $\bm{A}_{det}$.

If $\bm{A}_{det}$ includes two or more 1, then we calculate the CoG, $\bm{p}_G=[x_G,y_G,z_G]^{\rm{T}}$, of model $P_t$ using CAD. 
We determine posture $\hat{A}$ based on the CoG as follows:
\begin{equation}\forlab{cog}
\bm{p}_G = \frac{\Sigma_{i=0}^\eta {m_i} \bm{p}_{G,i}}{\Sigma_{i=0}^\eta {m_i}},
\end{equation}
where $m_i$ and $\bm{p}_{G,i}$ are the mass and CoG of the $i$-th part.

If $\bm{A}_{det}$ does not include two or more 1, we substitute the first element of $\bm{A}_{det}$ for $\hat{A}$.
Then, we set $\hat{P}$ to the bottom part in posture $\hat{A}$ and add $\hat{P}$ and $\hat{A}$ to $\hat{\bm{P}}^*$ and $\hat{\bm{A}}^*$. 
This one routine is repeated $\eta-1$ times where $\eta$ is the number of parts.

\section{Experiments}
\subsection{Outline}
We used the parts shown in~\figref{parts}(a), where a motor and a plate are fixed with bolts.
The parts were prepared for a belt drive unit used in an assembly challenge~\cite{Yokokohji2019}. 
Using the parts, we performed the following three experiments. 

The first experiment was to evaluate the proposed parts-fixing configuration algorithm.
Two cases of assembly sequences were tested, as described in~Section~\ref{eval-algo}. 

The second experiment evaluated two fixing abilities: an ability of maintaining the fixed pose and a holding ability against external forces.
We conducted parts-placement experiments with (Section~\ref{eval-val}) and without (Section~\ref{eval-force}) an external force application using a manipulator equipped with a parallel-jaw gripper. 
We evaluated the fixing ability based on holding forces, moving distances, and success rates.

The third experiment (Section~\ref{eval-feas}) involved verifying the feasibility of assembly operations (\figref{human-insertion}) with a robot arm equipped with a parallel-jaw gripper. 
%%%%%%%%%%%%%%%%%%%%%%%%%%%%%%%%%%
\begin{figure}[tb]
  \begin{minipage}[tb]{0.52\linewidth}
    \centering
    \includegraphics[width=0.9\linewidth]{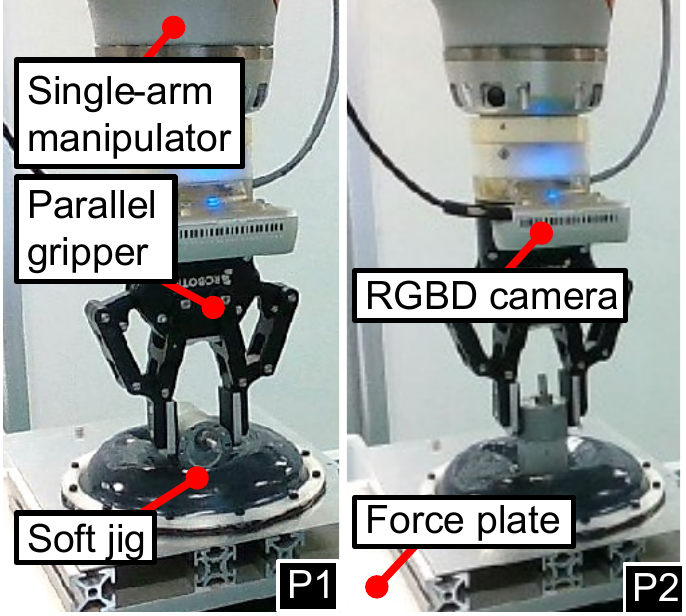}
    \subcaption{\small{Motor}}\figlab{place_motor}
  \end{minipage}
  \begin{minipage}[tb]{0.43\linewidth}
    \centering
    \includegraphics[width=0.95\linewidth]{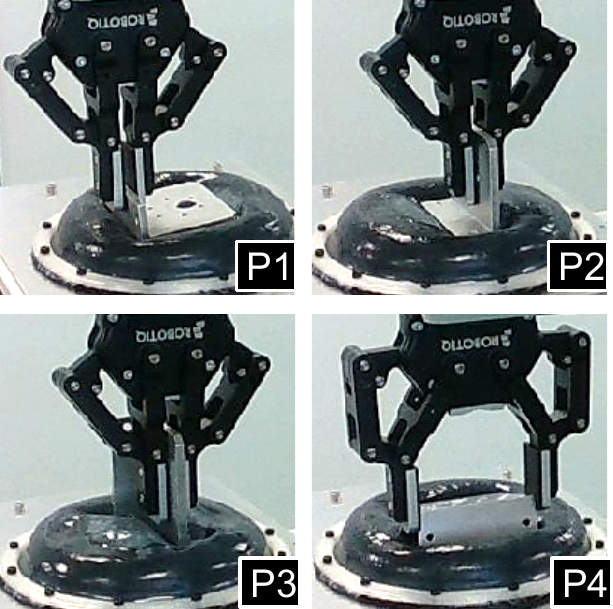}
    \subcaption{\small{Plate}}\figlab{place_plate}
  \end{minipage}
  \caption{\small{Two different postures of the motor and four different postures of the plate evaluated in the experiments described in Section~\ref{eval-val}.}}
  \figlab{pose-pattern-place}
\end{figure}
%%%%%%%%%%%%%%%%%%%%%%%%%%%%%%%%%%
\begin{figure}[tb]
  \begin{minipage}[tb]{\linewidth}
    \centering
    \includegraphics[width=0.65\linewidth]{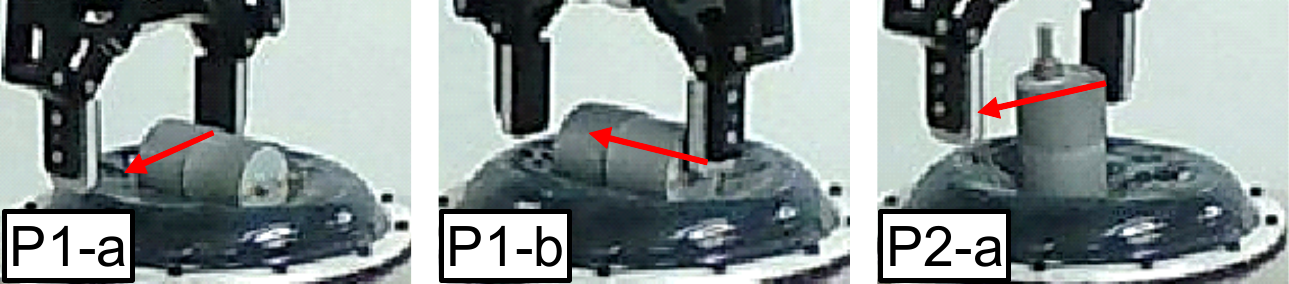}
    \subcaption{\small{Motor}}\figlab{force_motor}
  \end{minipage}
  \begin{minipage}[tb]{\linewidth}
    \centering
    \includegraphics[width=0.65\linewidth]{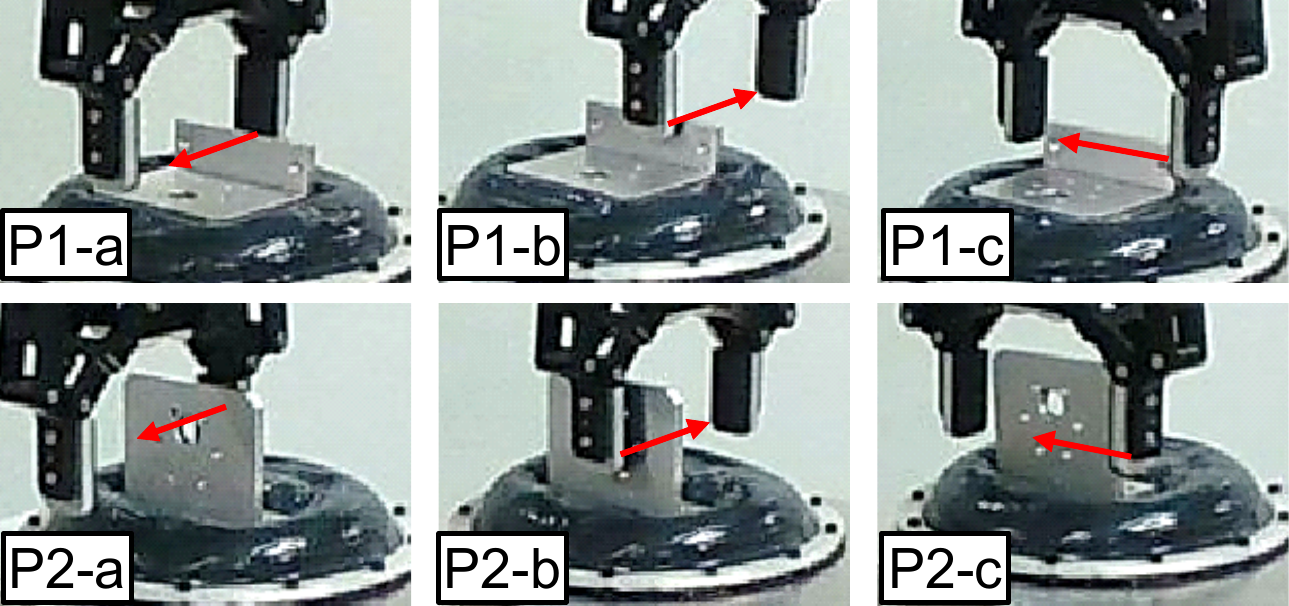}
    \subcaption{\small{Plate}}\figlab{force_plate}
  \end{minipage}
  \caption{\small{Different placement postures (P1 and P2) and pushing directions (a, b, and c) for the motor and plate evaluated in the experiments described in Section~\ref{eval-force}.}}
  \figlab{pose-pattern-force}
\end{figure}
%%%%%%%%%%%%%%%%%%%%%%%%%%%%%%%%%%
\begin{figure}[tb]
  \begin{minipage}[tb]{0.44\linewidth}
    \centering
    \includegraphics[width=0.70\linewidth]{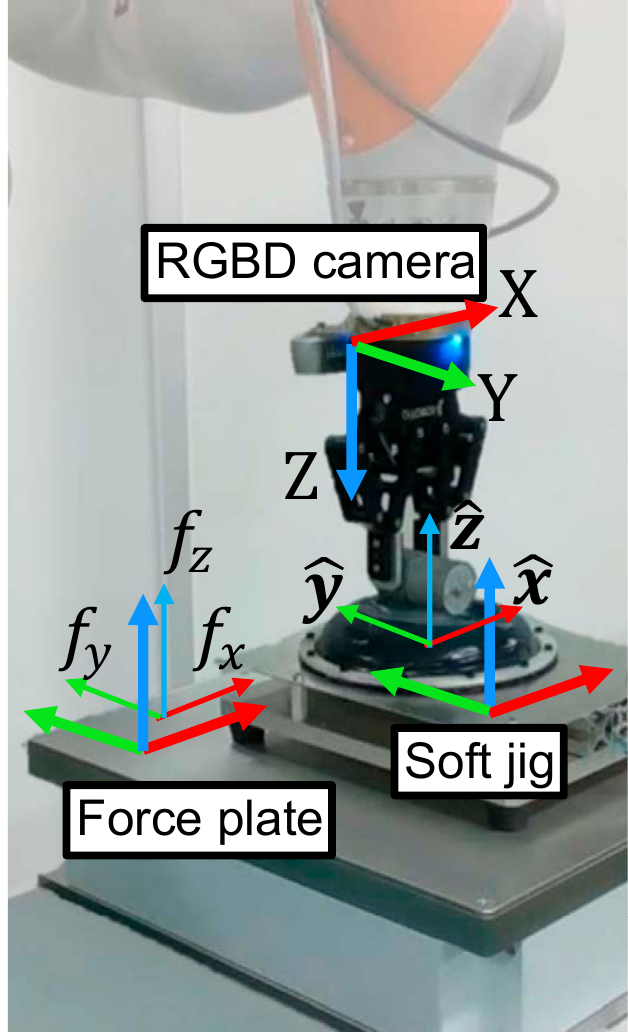}
    \subcaption{\small{Condition}}\figlab{before-ext-force}
  \end{minipage}
  \begin{minipage}[tb]{0.55\linewidth}
    \centering
    \includegraphics[width=0.69\linewidth]{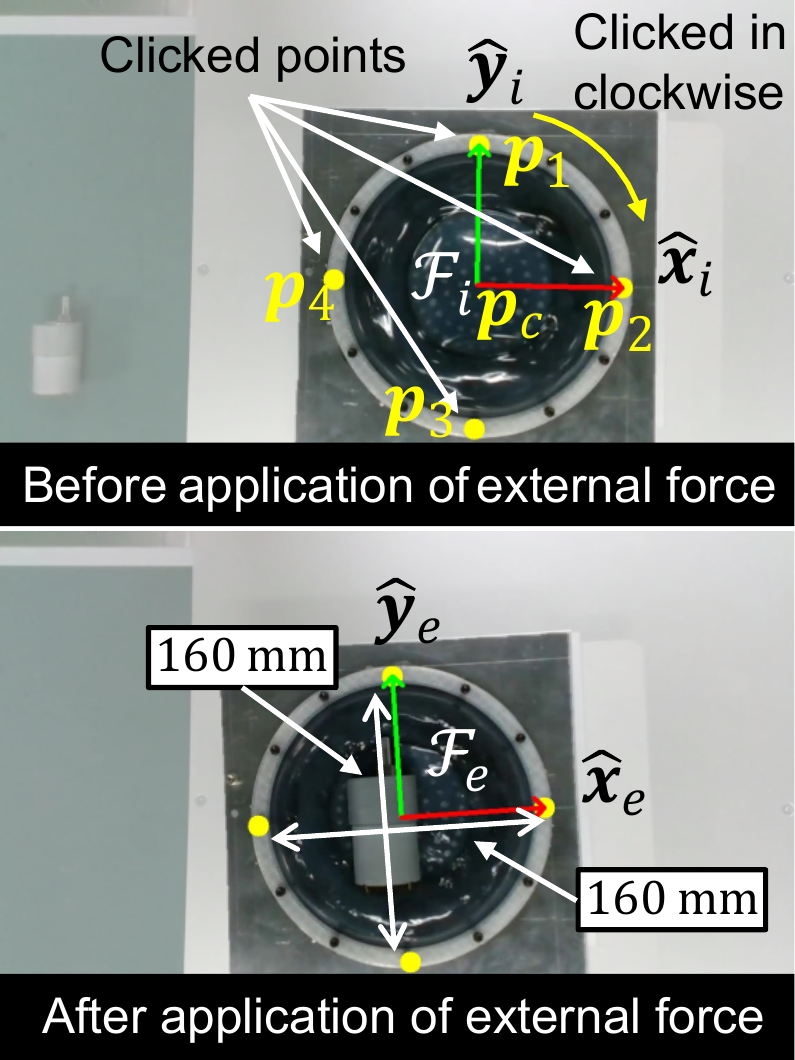}
    \subcaption{\small{Displacement}}\figlab{after-ext-force}
  \end{minipage}
  \caption{\small{Experiments to evaluate parts-fixing performance. We used a force plate to measure the normal and shearing forces. We calculated the displacement of the jig using the manually clicked points on both images before and after the application of the external forces.}}
  \figlab{exp-ext-force}
\end{figure}
%%%%%%%%%%%%%%%%%%%%%%%%%%%%%%%%%%
\begin{figure}[tb]
  \begin{minipage}[tb]{\linewidth}
    \centering
    \includegraphics[width=\linewidth]{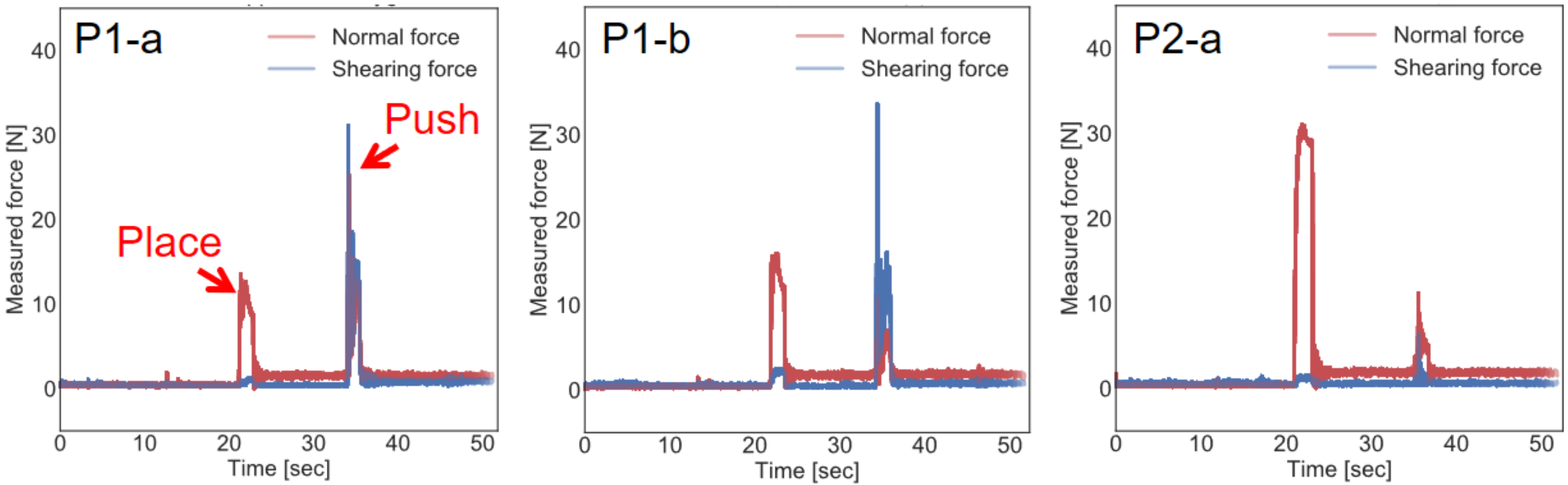}
    \subcaption{\small{Motor}}\figlab{plots_motor}
  \end{minipage}
  \begin{minipage}[tb]{\linewidth}
    \centering
    \includegraphics[width=\linewidth]{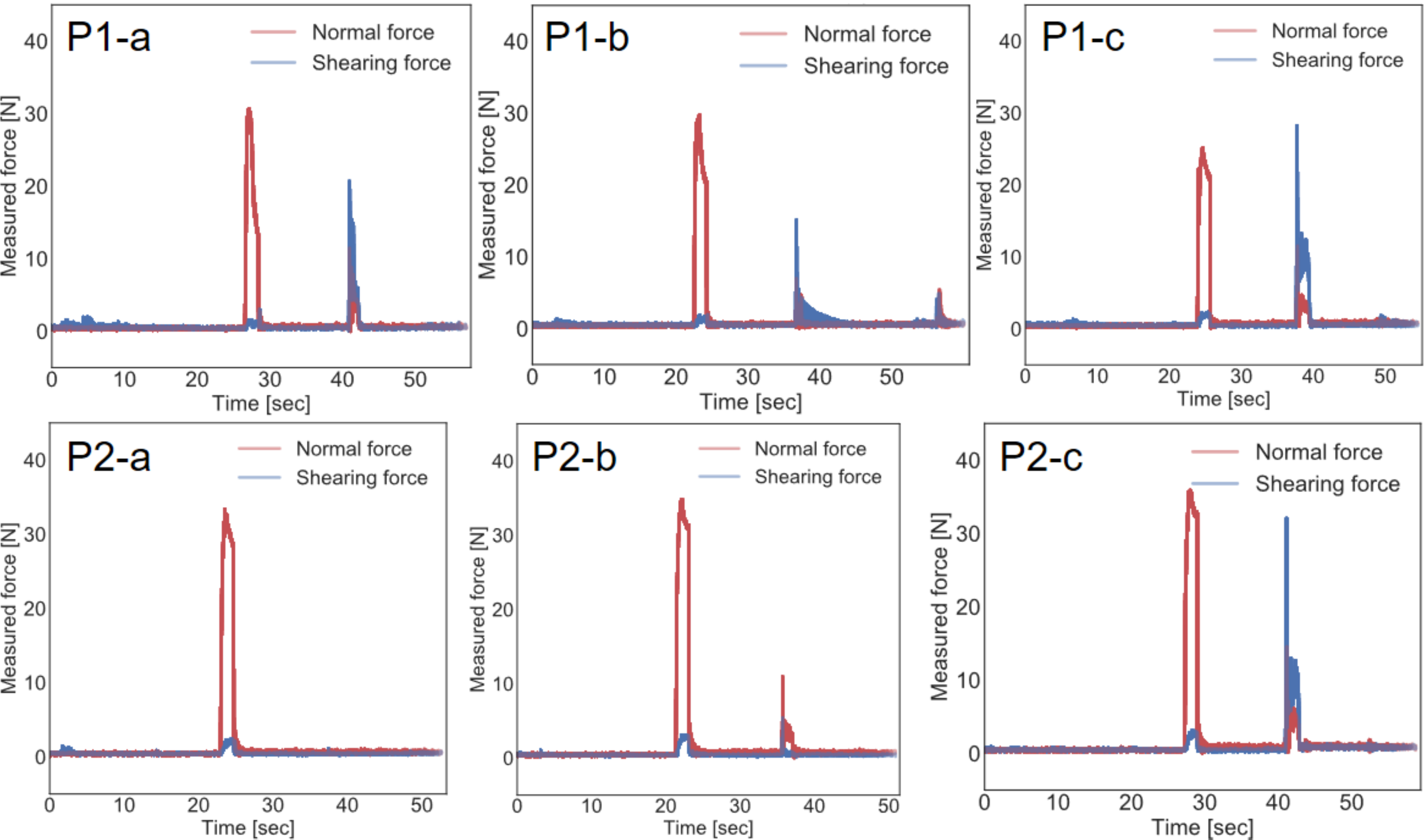}
    \subcaption{\small{Plate}}\figlab{plots_plate}
  \end{minipage}
  \caption{\small{Normal and shearing forces applied under the soft jig during the placing and pushing operations performed around the two peaks.}}
  \figlab{measured-force}
\end{figure}
%%%%%%%%%%%%%%%%%%%%%%%%%%%%%%%%%%
\begin{figure*}[tb]
  \begin{minipage}[tb]{0.38\linewidth}
    \centering
    \includegraphics[width=0.9\linewidth]{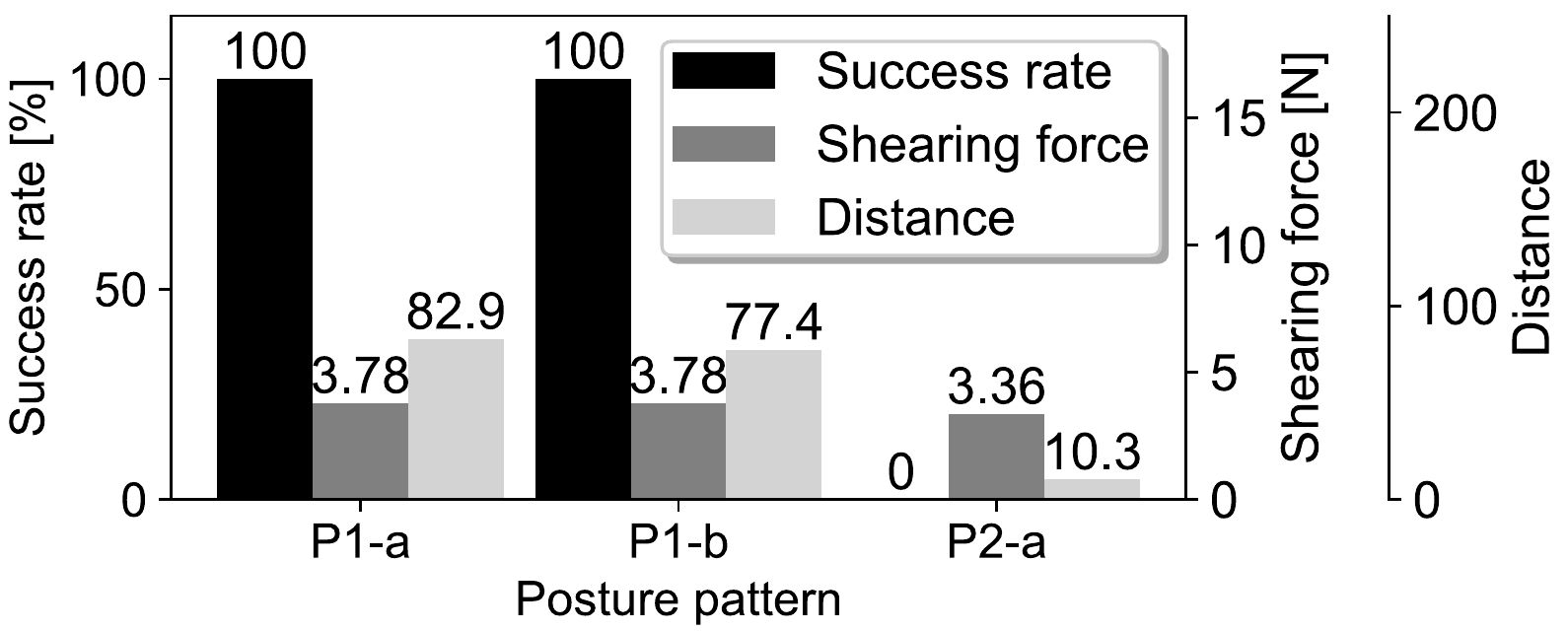}
    \subcaption{\small{Motor}}
  \end{minipage}
  \begin{minipage}[tb]{0.61\linewidth}
    \centering
    \includegraphics[width=0.9\linewidth]{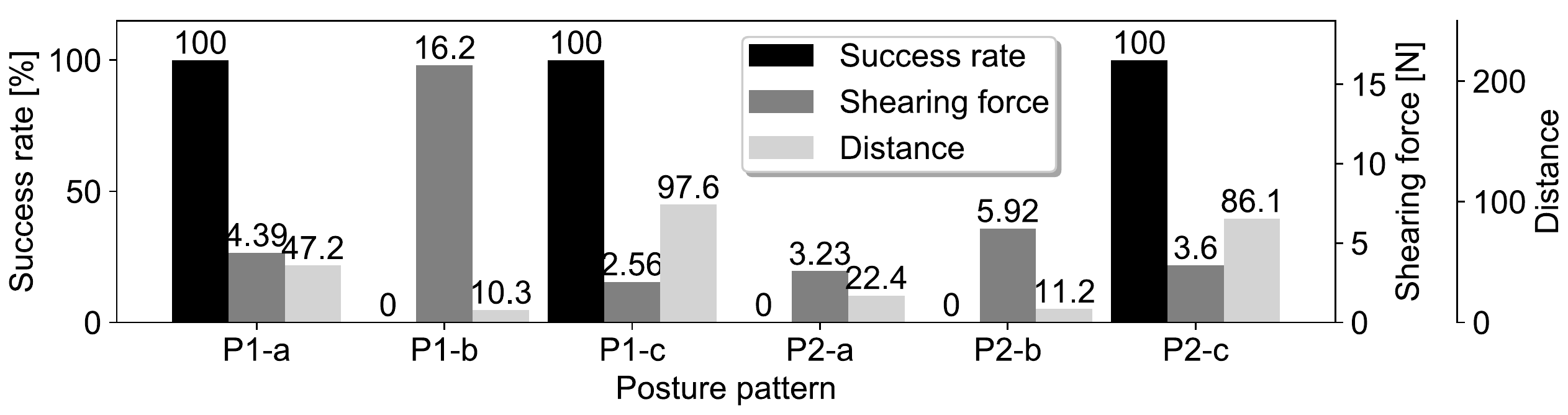}
    \subcaption{\small{Plate}}
  \end{minipage}
  \caption{\small{Performance of fixing the motor and plate. Each figure shows the success rate (number of successes in five trials), average of the maximum value of shearing force, and average distance of the soft jig before and after the external force application in five trials.}}\figlab{shape-force-rate}
\end{figure*}
%%%%%%%%%%%%%%%%%%%%%%%%%%%%%%%%%%
\begin{figure*}[tb]
    \centering
    \includegraphics[width=0.75\linewidth]{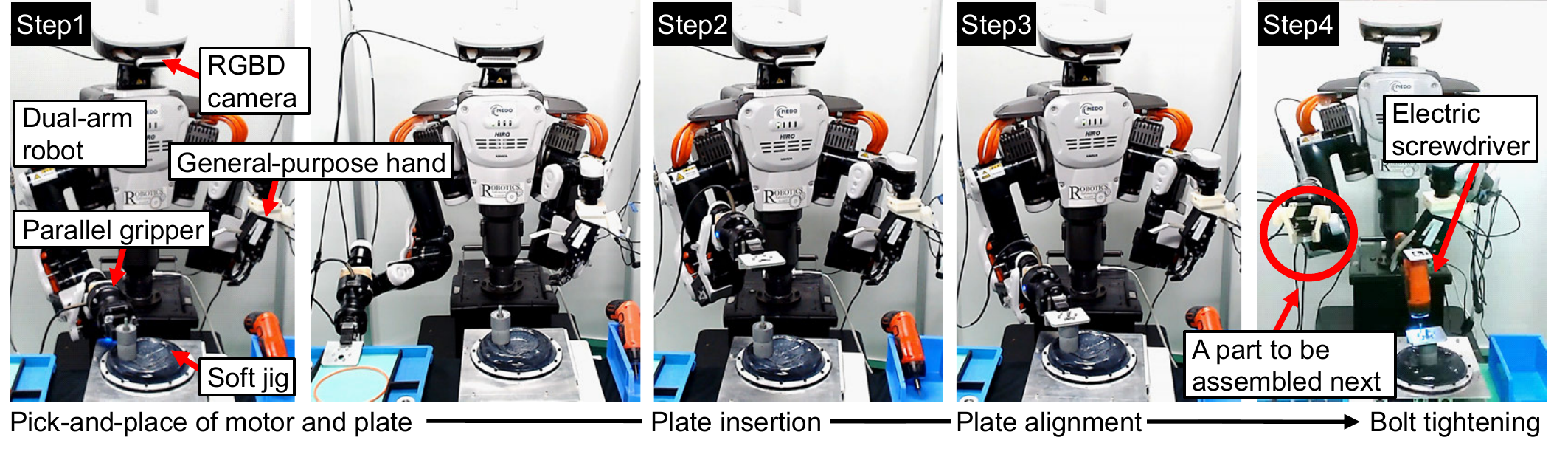}
    \caption{\small{Assembly sequence with the soft jig and a robot. The assembly order was the same as the manual assembly shown in~\figref{human-insertion}.}}
    \figlab{assembly-result}
\end{figure*}
%%%%%%%%%%%%%%%%%%%%%%%%%%%%%%%%%%

\subsection{Determining Fixed Parts and Postures} \label{eval-algo}
%%%%%%%%%%%%%%%%%%%%%%%%%%%%%%%%%%
\begin{table}[t]
    \centering
        \caption{\small{Fixing configurations determined by using the calculated values of \forref{place-determine}. Iterations 1 and 2 correspond to the iterations of the loop of~\algoref{fixing}.}}
        \tablab{reachable-direc}
        \begin{tabular}{p{3mm}p{18mm}p{13mm}p{18mm}p{13mm}} \toprule
            &\multicolumn{2}{c}{Iteration 1 $(i = 2)$}&\multicolumn{2}{c}{Iteration 2 $(i = 3)$}\\ 
            \multicolumn{1}{c}{Case}&\multicolumn{1}{c}{\forref{place-determine}}&\multicolumn{1}{c}{$\hat{A}$ / $\hat{P}$}& \multicolumn{1}{c}{\forref{place-determine}}&\multicolumn{1}{c}{$\hat{A}$ / $\hat{P}$}\\ \midrule
            \multicolumn{1}{c}{$1$}&$\bm{A}(\mathrm{motor,plate})$ &+z / motor& $\bm{A}(P_{c1}\mathrm{,bolts})^{\rm *a}$&+z / motor\rule[-1mm]{0mm}{4mm}\\
            \multicolumn{1}{c}{$2$}&$\bm{A}(\mathrm{plate,bolts})$ &+z / plate& $\bm{A}(P_{c2}\mathrm{,motor})^{\rm *b}$&-z / motor\rule[-1mm]{0mm}{4mm}\\
            \bottomrule
        \end{tabular}
        \begin{tablenotes}
        \item[a]\footnotesize{$^{\rm *a}$ $P_{c1}$ represents the combined part of motor and plate}
        \item[b]\footnotesize{$^{\rm *b}$ $P_{c2}$ represents the combined part of plate and bolts}
        \end{tablenotes}
\end{table}
%%%%%%%%%%%%%%%%%%%%%%%%%%%%%%%%%%
\tabref{reachable-direc} shows the results of the parts-fixing configuration for two cases of assembly sequences. Case 1 is \{motor, plate, bolts\}. Case 2 is \{plate, bolts, motor\}. 
The assembly of small bolts was not selected in the algorithm because if the bolts were positioned under the plate, the CoG position of the model combined with the plate and bolts was higher than that of the upside-down posture.

In Case 1, the parts-fixing configuration determined by \algoref{fixing} is the motor placed in the posture, as shown in Step~1 of~\figref{human-insertion}.
This posture is difficult to achieve in the base plate of a metal jig because pins are attached at the bottom of the motor, as shown in the top right image in Step~1 of~\figref{human-insertion}. 
This posture is achievable with the soft jig. 
The plate is inserted onto the fixed motor, and then, the bolts are screwed onto the plate fixed with the motor.

In Case 2, first, the bolts are placed onto the fixed plate in posture P4, as shown in~\figref{pose-pattern-place}(b).
Second, the plate with bolts is inserted onto the fixed motor.

\subsection{Evaluating Versatility to Fixed Parts and Postures} \label{eval-val}
To evaluate the versatility of the soft jig for the shape and posture of the parts, we investigated whether the fixed postures were maintained after they were released from the gripper. 

In addition to the determined posture, postures for comparisons were shown in~\figref{pose-pattern-place}. 
In the case of the motor, the shape is axisymmetric; therefore, there are two ways to place it in an axis-aligned manner.
The two postures are with the side (P1) or bottom surface (P2) of the cylinder shape being in contact with the jig. 

Four different postures of the plate were prepared.
These included the postures of the back or front side of the insertion hole facing straight up (\figref{pose-pattern-place}(b) P1 or P4).
We also prepared the postures where the bottom (P2) or side surface (P3) was in contact with the jig surface.

The gripper's trajectory and grasping configurations were manually generated. 
The trial was regarded as successful if the parts were upright, even if the gripper released the part, \ie~if the resting state was possible. 
We conducted 10 trials and visually checked whether each trial was a success or failure.
The success rates were 100\% ($= 10/10$) for all postures of the two parts; thus, the versatility of soft jig against the placed shapes is high.

\subsection{Evaluating Parts-Fixing Against External Force} \label{eval-force}
The fixing performance was evaluated based on the holding forces, moving distances, and success rates.
First, the fixing performance was evaluated based on the holding force when an external force was applied; the higher the performance, the higher the holding force should be.

The second was an evaluation based on the moving distances when an external force was applied.
We defined the fixing success based on the distance of the jig base itself, which was not fixed anywhere. 
The jig was moved in response to the pushing motion of \SI{70}{mm} straight-line trajectory to apply the external force.
If the part posture is not changed against the pushing motion, all the force should be converted into the jig movement, so the amount of movement is larger. 
If the posture is changed, that amount of the jig movement must be low.
Therefore, if the part is firmly fixed on the jig, it should move as much as the distance pushed by a robot; however, considering the elasticity of the silicon membrane, we defined a successful fixation as the distance more than \SI{63}{mm} (90\% of the \SI{70}{mm} pushing trajectory).

\subsubsection{Holding Force} \label{holding-force}
\figref{exp-ext-force}(a) shows the experimental setup including the robot arm with a gripper to apply the external force. 
We set a force plate and an RGBD camera to measure the holding force and the jig movement.
We measured the forces applied to the lower part of the jig when the fixed parts were pushed by the straight-line trajectory of the gripper. 
We also measured how much the soft jig moved before and after the external force application.
\figref{pose-pattern-force} shows the parts-postures in the experiments to evaluate the fixing performance.

Here, the resulting force on the contact surface was calculated as the magnitude of normal force $F_{n}$ and shearing force $F_{s}$ by using the following equations:
% \[
\begin{equation}\forlab{shearing_force}
    F_{n} = |f_{{z}}|,~~~
    F_{s} = \sqrt{f_{x}^2 + f_{y}^2},
\end{equation}
% \]
where $f_{x}$, $f_{y}$, and $f_{z}$ are the measured forces in the coordinate system of the force plate shown in~\figref{exp-ext-force}(a).
\figref{measured-force} shows the calculated values of the normal and shearing forces during the operations including placing and pushing of the parts on the jig. 
The graph IDs on the top left on each graph correspond to the IDs in~\figref{pose-pattern-force}. 
The two peaks on all graphs except P2-a of the plate show the force values at the timing of placement and pushing. 
In P2-a of the plate, the plate fell on the jig at the time when the gripper made contact with the plate in the pushing operation. 
Thus, only one peak existed because the forces by pushing could not be measured.

\subsubsection{Moving Distance} \label{moving-dist}
The displacements of the jig after the application of external force were also measured.
The hand-eye camera (Intel Corporation, RealSense D435) recorded RGBD images before and after applying the force to the part fixed on the jig in contact with the force plate.

We used two images shown in~\figref{exp-ext-force}(b) before and after applying the external force. 
We calculated the distance, $d(\mathcal{F}_{i},\mathcal{F}_{e})$, between the configuration frames of the soft jig before $\mathcal{F}_{i}$ and after $\mathcal{F}_{e}$ applying the external force:
\begin{equation}\forlab{distance_configs}
d(\mathcal{F}_{{i}},\mathcal{F}_{{e}}) \coloneqq \sqrt{\|\hat{\bm{x}}_{e}-\hat{\bm{x}}_{i}\|^2+\|\hat{\bm{y}}_{e}-\hat{\bm{y}}_{i}\|^2}.
\end{equation}
\forref{distance_configs} is a metric used to calculate the distance between the two configuration frames proposed in~\cite{Ahuactzin1999}. 
We calculated the poses ($\hat{\bm{x}}_{i}$, $\hat{\bm{y}}_{i}$) of $\mathcal{F}_{i}$ and the poses ($\hat{\bm{x}}_{e}$, $\hat{\bm{y}}_{e}$) of $\mathcal{F}_{e}$ in~[pixel] using four points, $\bm{p}_k (k=1,2,3,4) \in \mathbb{R} ^ 2$, as shown in \figref{exp-ext-force}(b). 
We clicked on the four screws of the jigs in the images as $\bm{p}_k$. 
Each screw had a central angle of $90^\circ$ and fixed the membrane onto the jig base. 
We converted the unit of the distance from pixels to mm based on the known width \SI{160}{mm} of the jig. $\hat{\bm{x}}$ and $\hat{\bm{y}}$ are calculated as
\begin{eqnarray}\forlab{point}
(\hat{\bm{x}},~\hat{\bm{y}}) &=& (\bm{p}_2-\bm{p}_{c},~\bm{p}_1-\bm{p}_{c}), \\
\bm{p}_{c} &=& \frac{\bm{p}_1+\bm{p}_2+\bm{p}_3+\bm{p}_4}{4}.
\end{eqnarray}

$\hat{\bm{z}}$ was set to $\bm{0}$ because the images shown in~\figref{exp-ext-force}(b) were captured from directly above; thus, the fixed surface of the jig remained horizontal even after the external force application.

\subsubsection{Factors Underlying Successful Parts-Fixing}
\figref{shape-force-rate} shows the average maximum values of the shearing force in five trials.
\figref{shape-force-rate} shows the success rates in the five trials. 
The success rate was 0\% for P2-a of Motor, P1-b of Plate, P2-a and P2-b. 
In these cases, the first peak of the normal force shown in~\figref{pose-pattern-force} indicates that the pushing force is applied to the same extent as in other cases. 
However, because the maximum force at the second peak of the shearing force is lower than other cases, the fixing fails.

\figref{shape-force-rate} shows the calculated values of $d(\mathcal{F}_{i},\mathcal{F}_{e})$. 
The failure cases resulted in a low amount of displacement compared to the successful ones. 
The difference between the mean displacements of the failure and success cases is \SI{24.0}{mm} $(= 78.2 - 54.2)$, and the value of the success cases is significantly larger than that of the failure cases. 
\SI{78.2}{mm} is larger than the original pushing distance \SI{70.0}{mm} because of over-displacement due to an acceleration of the pushing motion.
The jig is hardened by the jamming transition; thus, the deformation itself does not cause the small displacement.

In the case of low-height postures, such as P1-a and P1-b of the motor and P1-a and P1-c of the plate, as far as we visually confirmed, the displacement did not occur despite the direct external force. 
Thus, the success rates of the four cases were 100\%.
Posture P2-c of the plate was firmly fixed, although it was a high-height posture. 
This is because the pushing action of the external force pushes the fixed object into the inside of the jig as it tries to rotate on the axis perpendicular to the pushed direction.

Against the external forces, to fix high-height postures like P2-a of both parts, high datum planes surrounding the side surfaces of the parts are required as shown in Fig. 1 (d). 
Since such the large datum plane was not generated, the trials were unsuccessful.
The results suggest that selecting the placement posture and forming the datum plane are important.

\subsection{Feasibility of Assembly Operations for Fixed Parts} \label{eval-feas}
In this section, we confirm the force generated during the actual assembly operations by expanding the experiments described in the previous section.
The assembly operation by the dual-arm robot was executed by applying the procedure shown in~\figref{human-insertion}. 
Dual arms equipped two types of grippers: a parallel-jaw gripper used for grasping assembly parts and a general-purpose gripper used for grasping a tool such as the electric driver. 
As shown in~\figref{assembly-result}, the operation was divided into four steps. 
We used two arms to avoid regrasping a tool.
All operations were performed with handcrafted trajectories of one hand. 

During the experiments, given the assembly sequence and the gripper's trajectories, the robot could execute pick-and-place, insertion, and tightening of parts using the soft jig. 
The gripper did not apply external force directly to fixed parts; instead, the external force was applied via the grasped plate when the plate was inserted into the jig-fixed motor or when the grasped electric driver screwed off the bolts. 
Even under the external forces, the displacement of the motor on the jig was not significant, suggesting that it may be useful for fixing a part during assembly operations.

\section{Discussion}
%%%%%%%%%%%%%%%%%%%%%%%%%%%%%%%%%%
\begin{figure}[t]
    \centering
    \begin{minipage}[tb]{0.4\linewidth}
      \centering
      \includegraphics[width=0.88\linewidth]{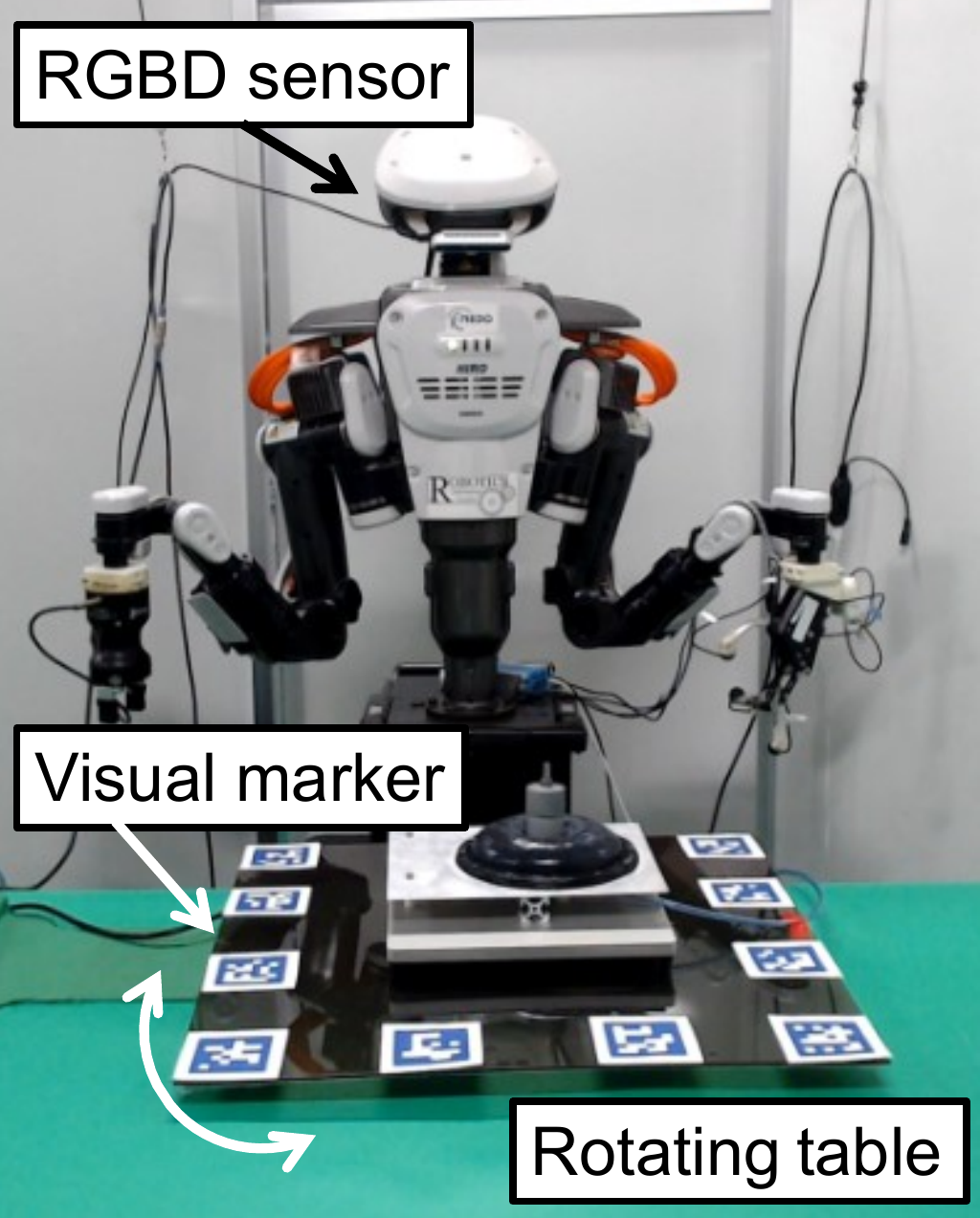}
      \subcaption{\small{System}}
    \end{minipage}
    \begin{minipage}[tb]{0.58\linewidth}
      \centering
      \includegraphics[width=0.9\linewidth]{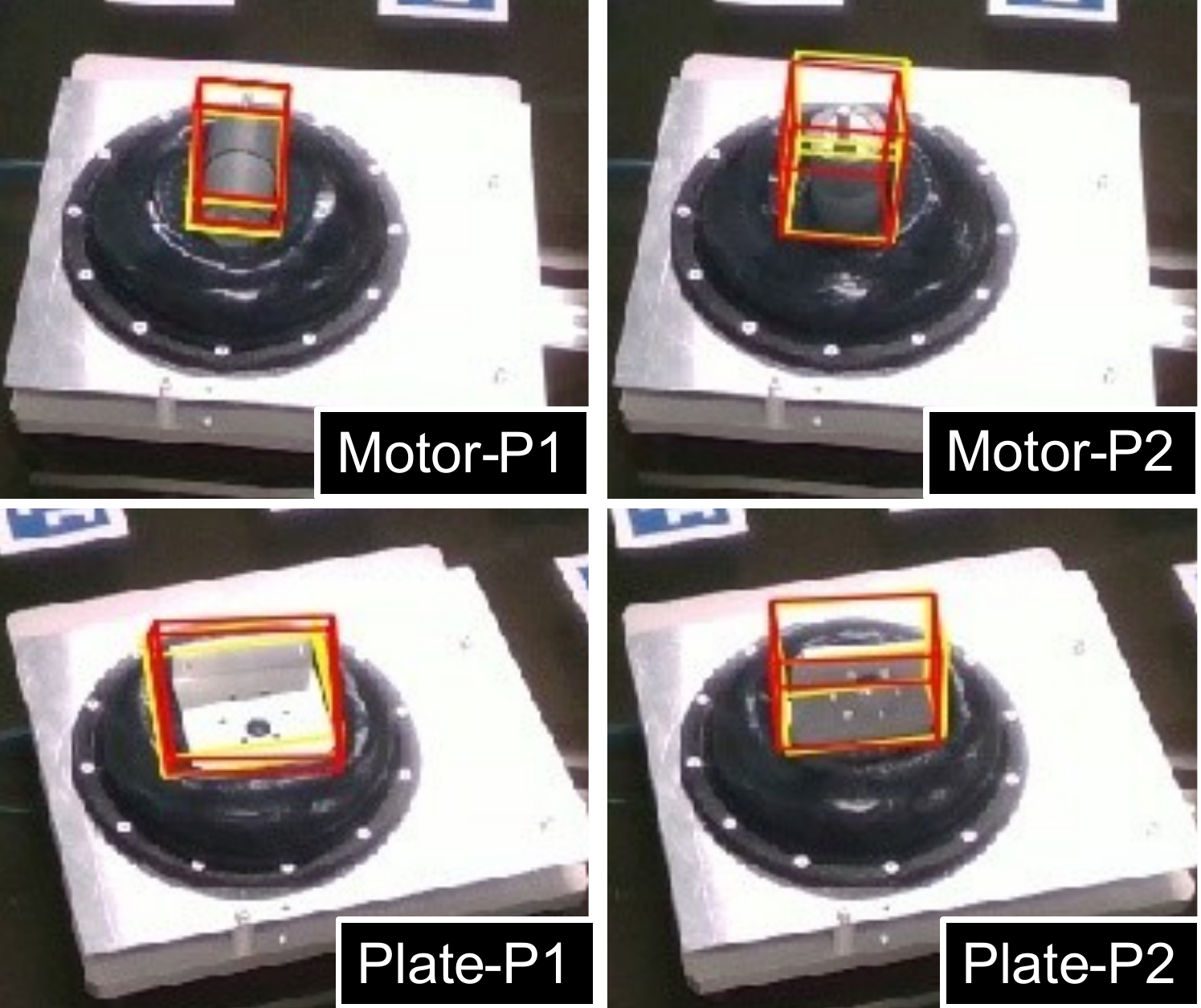}
      \subcaption{\small{Visualization of estimation results}}
    \end{minipage}
    \caption{\small{Dataset collection system and results of 6D pose estimation of the fixed parts.}}
    \figlab{pose-est}
\end{figure}
%%%%%%%%%%%%%%%%%%%%%%%%%%%%%%%%%%
To show the durability of the soft jig in industrial applications, we discuss the parts-pose estimation.
To design the concrete method is out-of-scope, but we discuss the possibility and future issues.

To confirm the 6D pose estimation task for fixed parts, we apply \textit{PVNet}~\cite{Peng2019}, one of deep learning-based algorithms~\cite{Tekin2018,Tremblay2018}.
To train the network, we leverage a quick dataset collection method using visual markers proposed in~\cite{Hinterstoisser2013,KiyokawaRAL2019,KiyokawaAR2019}, as shown in~\figref{pose-est}(a). 
These methods reduce the human effort required for training and enable the use of PVNet for the high-mix low-volume production.

\figref{pose-est} (b) shows the results of the pose estimation applied to test images showing different poses of the motor and the plate. 
The yellow 3D bounding boxes in the images show the box calculated based on the visual markers, which is the ground-truth.
The red boxes with similar shapes to the yellow boxes show the 6D poses estimated by the trained model.
The results suggest that the pose in a wide viewing angle can be accurately tracked, and the appearance of the soft jig does not deteriorate the pose estimation.

\section{Conclusion}
To design a general-purpose assembly robot system, we proposed a soft jig, a deformable fixture that replaces a custom-made metal jig. 
To leverage the soft jig efficiently for assemblies, we proposed a method to configure parts-fixing based on the contact relations, reachable directions, and center of gravity of the fixed parts.

We evaluated the following four factors: usability of the parts-fixing configuration algorithm using two assembly sequences, ability to maintain the fixed postures, holding ability against external forces, and applicability for inserting and bolting parts in assemblies by an actual robot.

The proposed algorithm could determine fixed parts and their poses for two assembly sequences of a product with parts of various shapes, including a motor, a plate, and bolts.
In this study, parts-placement in various fixing-postures was successful in all cases.
Successful parts-fixing was demonstrated in cases of where fixing occured in postures with low CoG.
Through the experiments, we confirmed that selecting the placement posture and forming the datum plane are important for firm fixing.

\bibliographystyle{IEEEtran}
\footnotesize
\bibliography{bibliography/reference}

\begin{thebibliography}{10}
\providecommand{\url}[1]{#1}
\csname url@rmstyle\endcsname
\providecommand{\newblock}{\relax}
\providecommand{\bibinfo}[2]{#2}
\providecommand\BIBentrySTDinterwordspacing{\spaceskip=0pt\relax}
\providecommand\BIBentryALTinterwordstretchfactor{4}
\providecommand\BIBentryALTinterwordspacing{\spaceskip=\fontdimen2\font plus
\BIBentryALTinterwordstretchfactor\fontdimen3\font minus
  \fontdimen4\font\relax}
\providecommand\BIBforeignlanguage[2]{{%
\expandafter\ifx\csname l@#1\endcsname\relax
\typeout{** WARNING: IEEEtran.bst: No hyphenation pattern has been}%
\typeout{** loaded for the language `#1'. Using the pattern for}%
\typeout{** the default language instead.}%
\else
\language=\csname l@#1\endcsname
\fi
#2}}

\bibitem{Gunasekaran1999}
A.~Gunasekaran, ``Agile manufacturing: A framework for research and
  development,'' \emph{Int. J. of Production Economics}, vol.~62, no.~1, pp.
  87--105, 1999.

\bibitem{Costa2017}
R.~J. Costa, F.~Silva, and R.~D. Campilho, ``A novel concept of agile assembly
  machine for sets applied in the automotive industry,'' \emph{The Int. J. of
  Adv. Manuf. Technology}, vol.~91, pp. 4043--4054, 2017.

\bibitem{Maeda2003}
Y.~Maeda, H.~Kikuchi, H.~Izawa, H.~Ogawa, M.~Sugi, and T.~Arai, ``An easily
  reconfigurable robotic assembly system,'' in \emph{ICRA}, 2003, pp.
  2586--2591.

\bibitem{Kim2019}
W.~{Kim}, M.~{Lorenzini}, P.~{Balatti}, P.~D.~H. {Nguyen}, U.~{Pattacini},
  V.~{Tikhanoff}, L.~{Peternel}, C.~{Fantacci}, L.~{Natale}, G.~{Metta}, and
  A.~{Ajoudani}, ``Adaptable workstations for human-robot collaboration: A
  reconfigurable framework for improving worker ergonomics and productivity,''
  \emph{IEEE RAM}, vol.~26, no.~3, pp. 14--26, 2019.

\bibitem{Gorjup2020}
G.~Gorjup, G.~Gao, A.~Dwivedi, and M.~Liarokapis, ``Combining compliance
  control, {CAD} based localization, and a multi-modal gripper for rapid and
  robust programming of assembly tasks,'' in \emph{IROS}, 2020, pp. 9064--9071.

\bibitem{Rajan1999}
V.~N. Rajan, K.~Sivasubramanian, and J.~E. Fernandez, ``Accessibility and
  ergonomic analysis of assembly product and jig designs,'' \emph{International
  Journal of Industrial Ergonomics}, vol.~23, no.~5, pp. 473--487, 1999.

\bibitem{Zhang2019}
H.~Zhang, L.~Zheng, P.~Wang, and W.~Fan, ``Intelligent configuring for agile
  joint jig based on smart composite jig model,'' \emph{The Int. J. of Adv.
  Manuf. Technology}, vol. 105, pp. 3927–--3949, 2019.

\bibitem{Trappey2005}
J.~C. Trappey and C.~R. Liu, ``A literature survey of fixturedesign
  automation,'' \emph{Int. J. of Adv. Manuf. Technology}, vol.~5, pp.
  240--–255, 1990.

\bibitem{Naing2000}
S.~Naing, G.~Burley, R.~Odi, A.~Williamson, and J.~Corbett, ``Design for
  tooling to enable jigless assembly – an integrated methodology for jigless
  assembly,'' \emph{SAE Trans.}, vol. 109, pp. 299--311, 2000.

\bibitem{Kim2017}
Y.-L. Kim, H.-C. Song, and J.-B. Song, ``Force control based jigless assembly
  strategy of a unit box using dual-arm and friction,'' in \emph{ISR}, 2013,
  pp. 1--3.

\bibitem{Bi2001}
Z.~M. Bi and W.~J. Zhang, ``Flexible fixture design and automation: Review,
  issues and future directions,'' \emph{Int. J. of Production Research},
  vol.~39, no.~13, pp. 2867--2894, 2001.

\bibitem{Grippo1987}
P.~M. Grippo, M.~V. Gandhi, and B.~S. Thompson, ``The computer-aided design of
  modular fixturing systems,'' \emph{The Int. J. of Adv. Manuf. Technology},
  vol.~2, no.~2, pp. 75--88, 1987.

\bibitem{Whybrew1992}
K.~Whybrew and B.~K.~A. Ngoi, ``Computer aided design of modular fixture
  assembly,'' \emph{The Int. J. of Adv. Manuf. Technology}, vol.~7, pp.
  267--276, 1992.

\bibitem{Fathianathan2007}
M.~Fathianathan, A.~S. Kumar, and A.~Y.~C. Nee, ``An adaptive machining fixture
  design system for automatically dealing with design changes,'' \emph{ASME. J.
  Comput. Inf. Sci. Eng.}, vol.~7, no.~3, pp. 259--268, 2007.

\bibitem{Shi2020}
P.~Shi, Z.~Hu, K.~Nagata, W.~Wan, Y.~Domae, and K.~Harada, ``An adaptive pin
  array fixture to fix multiple parts,'' in \emph{ROBOMECH}, 2020, pp.
  2A2--K10.

\bibitem{Lee2005}
C.~Lee, M.~Kim, Y.~J. Kim, N.~Hong, S.~Ryu, H.~J. Kim, and S.~Kim, ``Soft robot
  review,'' \emph{Int. J. of Control, Automation and Systems}, vol.~15, pp.
  3--15, 2017.

\bibitem{Watanabe2017}
T.~Watanabe, K.~Yamazaki, and Y.~Yokokohji, ``Survey of robotic manipulation
  studies intending practical applications in real environments -object
  recognition, soft robot hand, and challenge program and benchmarking-,''
  \emph{Advanced Robotics}, vol.~31, no. 19--20, pp. 1114--1132, 2017.

\bibitem{Brown2010}
E.~Brown, N.~Rodenberg, J.~Amend, A.~Mozeika, E.~Steltz, M.~R. Zakin,
  H.~Lipson, and H.~M. Jaeger, ``Universal robotic gripper based on the jamming
  of granular material,'' \emph{PNAS}, vol. 107, no.~44, pp. 18\,809--18\,814,
  2010.

\bibitem{Sakuma2018}
T.~Sakuma, F.~von Drigalski, M.~Ding, J.~Takamatsu, and T.~Ogasawara, ``A
  universal gripper using optical sensing to acquire tactile information and
  membrane deformation,'' in \emph{IROS}, 2018, pp. 6431--6436.

\bibitem{Alspach2019}
A.~Alspach, K.~Hashimoto, N.~Kuppuswamy, and R.~Tedrake, ``Soft-bubble: A
  highly compliant dense geometry tactile sensor for robot manipulation,'' in
  \emph{RoboSoft}, 2019, pp. 597--604.

\bibitem{Sakuma2019}
T.~Sakuma, E.~Phillips, G.~A.~G. Ricardez, M.~Ding, J.~Takamatsu, and
  T.~Ogasawara, ``A parallel gripper with a universal fingertip device using
  optical sensing and jamming transition for maintaining stable grasps,'' in
  \emph{IROS}, 2019, pp. 5814--5819.

\bibitem{Lu2020}
Q.~{Lu}, L.~{He}, T.~{Nanayakkara}, and N.~{Rojas}, ``Precise in-hand
  manipulation of soft objects using soft fingertips with tactile sensing and
  active deformation,'' in \emph{RoboSoft}, 2020, pp. 52--57.

\bibitem{Lathrop2020}
E.~{Lathrop}, I.~{Adibnazari}, N.~{Gravish}, and M.~T. {Tolley}, ``Shear
  strengthened granular jamming feet for improved performance over natural
  terrain,'' in \emph{RoboSoft}, 2020, pp. 388--393.

\bibitem{Chopra2020}
S.~{Chopra}, M.~T. {Tolley}, and N.~{Gravish}, ``Granular jamming feet enable
  improved foot-ground interactions for robot mobility on deformable ground,''
  \emph{IEEE RA-L}, vol.~5, no.~3, pp. 3975--3981, 2020.

\bibitem{Fujita2018}
M.~Fujita, S.~Ikeda, T.~Fujimoto, T.~Shimizu, S.~Ikemoto, and T.~Miyamoto,
  ``Development of universal vacuum gripper for wall-climbing robot,''
  \emph{Advanced Robotics}, vol.~32, no.~6, pp. 283--296, 2018.

\bibitem{HamayaIROS}
M.~{Hamaya}, F.~{von Drigalski}, T.~{Matsubara}, K.~{Tanaka}, R.~{Lee},
  C.~{Nakashima}, Y.~{Shibata}, and Y.~{Ijiri}, ``Learning soft robotic
  assembly strategies from successful and failed demonstrations,'' in
  \emph{IROS}, 2020, pp. 8309--8315.

\bibitem{HamayaICRA}
M.~{Hamaya}, R.~{Lee}, K.~{Tanaka}, F.~{von Drigalski}, C.~{Nakashima},
  Y.~{Shibata}, and Y.~{Ijiri}, ``Learning robotic assembly tasks with lower
  dimensional systems by leveraging physical softness and environmental
  constraints,'' in \emph{ICRA}, 2020, pp. 7747--7753.

\bibitem{Yamazaki2018}
K.~Yamazaki, T.~Higashide, D.~Tanaka, and K.~Nagahama, ``Assembly manipulation
  understanding based on {3D} object pose estimation and human motion
  estimation,'' in \emph{ROBIO}, 2018, pp. 802--807.

\bibitem{Fukuda2019}
K.~Fukuda, I.~G. Ramirez-Alpizar, N.~Yamanobe, D.~Petit, K.~Nagata, and
  K.~Harada, ``Recognition of assembly tasks based on the actions associated to
  the manipulated objects,'' in \emph{SII}, 2019, pp. 193--198.

\bibitem{Amend2012}
J.~R. {Amend}, E.~{Brown}, N.~{Rodenberg}, H.~M. {Jaeger}, and H.~{Lipson}, ``A
  positive pressure universal gripper based on the jamming of granular
  material,'' \emph{IEEE Trans. on Robotics}, vol.~28, no.~2, pp. 341--350,
  2012.

\bibitem{Bedeoui2019}
A.~Bedeoui, R.~Ben~Hadj, M.~Hammadi, M.~Trigui, and N.~Aifaoui, ``Assembly
  sequence plan generation of heavy machines based on the stability
  criterion,'' \emph{The Int. J. of Adv. Manuf. Technology}, vol. 102, pp.
  2745--2755, 2019.

\bibitem{Tariki2020}
K.~Tariki, T.~Kiyokawa, G.~A.~G. Ricardez, J.~Takamatsu, and T.~Ogasawara,
  ``{3D} model-based assembly sequence optimization using insertionable
  properties of parts,'' in \emph{SII}, 2020, pp. 1400--1405.

\bibitem{TarikiAR}
K.~Tariki, T.~Kiyokawa, T.~Nagatani, J.~Takamatsu, and T.~Ogasawara,
  ``Generating complex assembly sequences from {3D} {CAD} models considering
  insertion relations,'' \emph{Advanced Robotics}, vol.~35, no.~6, pp.
  337--348, 2020.

\bibitem{Yokokohji2019}
Y.~Yokokohji, Y.~Kawai, M.~Shibata, Y.~Aiyama, S.~Kotosaka, W.~Uemura, A.~Noda,
  H.~Dobashi, T.~Sakaguchi, and K.~Yokoi, ``{Assembly Challenge: a robot
  competition of the Industrial Robotics Category, World Robot Summit –
  summary of the pre-competition in 2018},'' \emph{Advanced Robotics}, vol.~33,
  no.~17, pp. 876--899, 2019.

\bibitem{Ahuactzin1999}
J.-M. Ahuactzin and K.~Gupta, ``The kinematic roadmap: A motion planning based
  global approach for inverse kinematics of redundant robots,'' \emph{IEEE
  Trans. on Robotics and Automation}, vol.~15, no.~4, pp. 653--669, 1999.

\bibitem{Peng2019}
S.~Peng, Y.~Liu, Q.~Huang, X.~Zhou, and H.~Bao, ``{PVNet}: Pixel-wise voting
  network for {6DoF} pose estimation,'' in \emph{CVPR}, 2019, pp. 4561--4570.

\bibitem{Tekin2018}
B.~Tekin, S.~N. Sinha, and P.~Fua, ``Real-time seamless single shot {6D} object
  pose prediction,'' in \emph{CVPR}, 2018, pp. 292--301.

\bibitem{Tremblay2018}
J.~Tremblay, T.~To, B.~Sundaralingam, Y.~Xiang, D.~Fox, and S.~Birchfield,
  ``Deep object pose estimation for semantic robotic grasping of household
  objects,'' in \emph{CoRL}, vol.~87, 2018, pp. 306--316.

\bibitem{Hinterstoisser2013}
S.~Hinterstoisser, V.~Lepetit, S.~Ilic, S.~Holzer, G.~Bradski, K.~Konolige, and
  N.~Navab, ``Model based training, detection and pose estimation of
  texture-less {3D} objects in heavily cluttered scenes,'' in \emph{ACCV},
  2012, pp. 548--562.

\bibitem{KiyokawaRAL2019}
T.~Kiyokawa, K.~Tomochika, J.~Takamatsu, and T.~Ogasawara, ``Fully automated
  annotation with noise-masked visual markers for deep-learning-based object
  detection,'' \emph{IEEE RA-L}, vol.~4, no.~2, pp. 1972--1977, 2019.

\bibitem{KiyokawaAR2019}
{T. Kiyokawa, K. Tomochika, J. Takamatsu, and T. Ogasawara}, ``Efficient
  collection and automatic annotation of real-world object images by taking
  advantage of post-diminished multiple visual markers,'' \emph{Advanced
  Robotics}, vol.~33, no.~24, pp. 1264--1280, 2019.

\end{thebibliography}
\end{document}